\documentclass[10pt,twocolumn,letterpaper]{article}

\usepackage{cvpr}
\usepackage{times}
\usepackage{epsfig}
\usepackage{graphicx}
\usepackage{amsmath}
\usepackage{amssymb}
\usepackage{epstopdf}
\usepackage{algorithm}
\usepackage{algorithmic}
\usepackage{multirow}
\usepackage{array}
\usepackage{booktabs}
\usepackage{subfigure}
\usepackage{color}
\usepackage{xcolor}

\usepackage[breaklinks=true,bookmarks=false]{hyperref}

\cvprfinalcopy 

\def\cvprPaperID{****} 

\begin{document}

\title{Instance-Invariant Domain Adaptive Object Detection via Progressive Disentanglement}

\author{Aming Wu$^1$\thanks{This work was done when Aming Wu visited ReLER Lab, UTS.}, \; Yahong Han$^1$, \; Linchao Zhu$^2$, \; Yi Yang$^2$\\
$^1$College of Intelligence and Computing, Tianjin University, Tianjin, China\\
$^2$ReLER, University of Technology Sydney, Australia\\
{\tt\small \{tjwam,yahong\}@tju.edu.cn}, {\tt\small \{Linchao.Zhu,yi.yang\}@uts.edu.au}
}

\maketitle

\begin{abstract}
    Most state-of-the-art methods of object detection suffer from poor generalization ability when the training and test data are from different domains, e.g., with different styles. To address this problem, previous methods mainly use holistic representations to align feature-level and pixel-level distributions of different domains, which may neglect the instance-level characteristics of objects in images. Besides, when transferring detection ability across different domains, it is important to obtain the instance-level features that are domain-invariant, instead of the styles that are domain-specific. Therefore, in order to extract instance-invariant features, we should disentangle the domain-invariant features from the domain-specific features. To this end, a progressive disentangled framework is first proposed to solve domain adaptive object detection. Particularly, base on disentangled learning used for feature decomposition, we devise two disentangled layers to decompose domain-invariant and domain-specific features. And the instance-invariant features are extracted based on the domain-invariant features. Finally, to enhance the disentanglement, a three-stage training mechanism including multiple loss functions is devised to optimize our model. In the experiment, we verify the effectiveness of our method on three domain-shift scenes. Our method is separately 2.3\%, 3.6\%, and 4.0\% higher than the baseline method \cite{saito2019strong}.
\end{abstract}

\vspace{-0.2in}
\section{Introduction}

Recently, great efforts have been made on object detection \cite{girshick2015fast,ren2015faster,hu2018relation,lin2017focal,redmon2016you}. Though most state-of-the-art methods achieve outstanding detection performance on many benchmarks \cite{everingham2010pascal,lin2014microsoft}, they suffer from poor generalization ability when the training and test images are from different domains, which is cast into the setting of domain adaptive object detection (DAOD). In the task of DAOD, domain gap always exists between the source/training and target/test images, e.g., with different illuminations and different styles etc. Although the performance could be improved via collecting additional images with well-labeled objects from the target domain, it is time-consuming and labor-intensive.

\begin{figure}
  \centering
  \begin{minipage}[t]{1.0\linewidth}
    \includegraphics[width=1.0\linewidth]{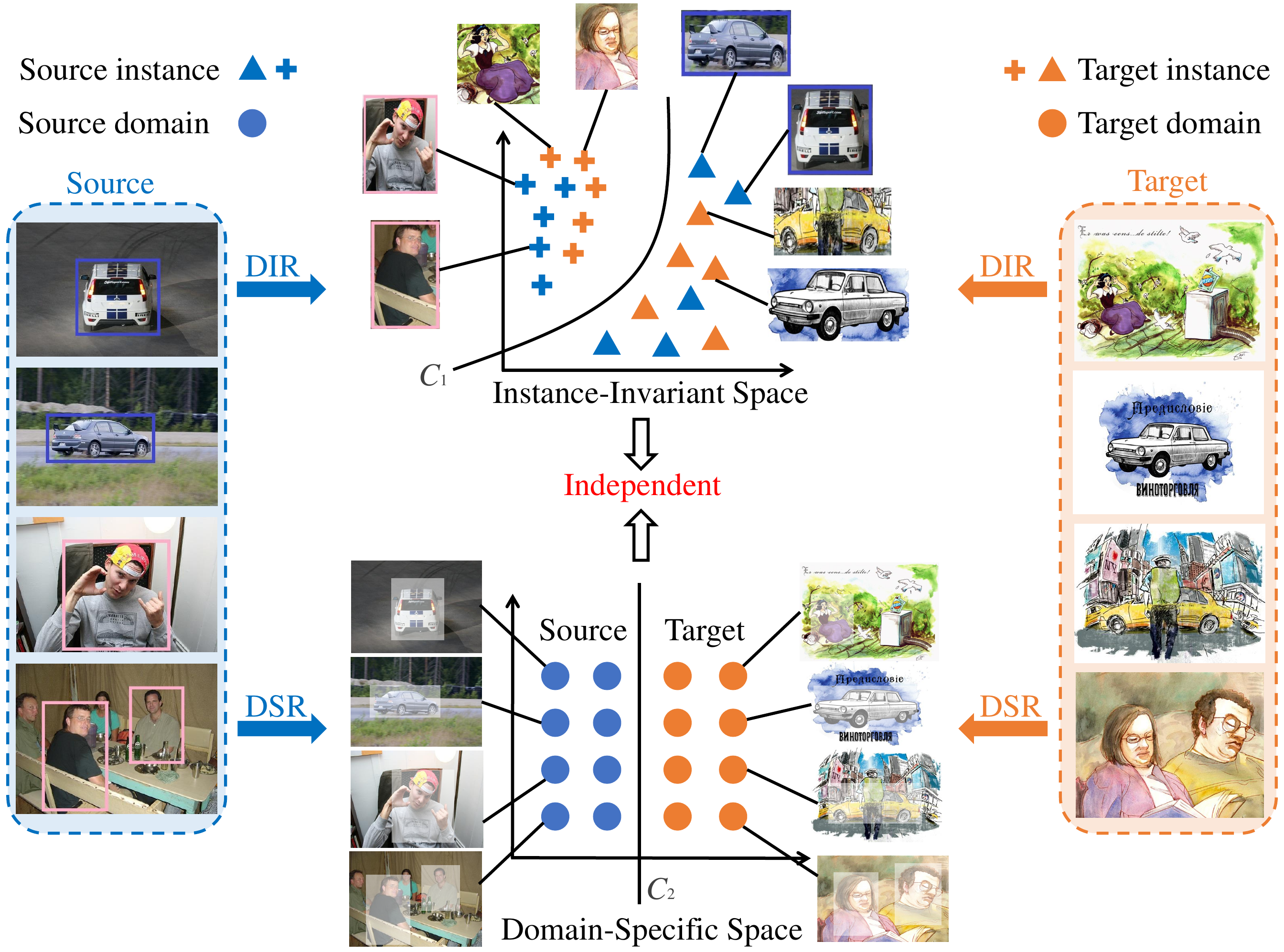}
  \end{minipage}
  \caption{The process of our disentangled method for domain adaptive object detection. We decompose source and target image representations into domain-invariant representations (DIR) and domain-specific representations (DSR). Then, we extract from DIR the instance-invariant representations that lie in an instance-invariant space, in which the instance-invariant features are used to describe the characteristics of objects. In the instance-invariant space, we conduct instance classification (i.e., via ${C}_{\rm 1}$) for the adaptive object detection. And different domains could be easily distinguished (i.e., via ${C}_{\rm 2}$) in the domain-specific space.}
  \label{fig1}
  \vspace{-0.2in}
\end{figure}

In order to alleviate the impact of domain-shift \cite{ganin2014unsupervised}, representative methods \cite{chen2018domain,saito2019strong,he2019multi} towards DAOD employ unsupervised domain adaptation \cite{roy2019unsupervised,pan2019transferrable,zhang2019domain} to align distributions of different domains, e.g., via adversarial training \cite{ganin2014unsupervised} or style translation \cite{kim2019diversify}. Distribution alignment is always conducted in a holistic representation (e.g., in feature-level \cite{cicek2019unsupervised,lee2019sliced} or pixel-level \cite{gong2019dlow,bousmalis2017unsupervised,russo2018source}) of source and target images, which may neglect the instance-level characteristics of objects in images, such as object locations or basic shapes of objects etc. When transferring detection ability from source images to target images, it is the instance-level features that really count, which are always domain-invariant, not the illuminations and painting styles that are domain-specific. Therefore, in order to obtain the instance-invariant features and bridge the domain gap in DAOD, we should try to disentangle the domain-invariant representations (DIR) from the domain-specific representations (DSR).

As a method of feature decomposition, disentangled learning \cite{do2019theory,locatello2018challenging} has been demonstrated to be effective in tasks of few-shot learning \cite{ridgeway2018learning,scott2018adapted} and image translation \cite{lee2018diverse,huang2018multimodal}. The purpose of disentangled learning is to uncover a set of independent factors that give rise to the current observation \cite{do2019theory}. And the major advantage is that disentangled representations could contain all the information presents in the current observation in a compact and interpretable structure while being independent of the current task \cite{locatello2018challenging,bengio2013representation}. In this paper, we propose to employ disentangled learning to disentangle an image representation into a domain-invariant representation (DIR) and a domain-specific representation (DSR) (see Fig. \ref{fig1}), so as to obtain the instance-invariant representation (IIR). Taking the IIR as a bridge, we have great potential to strengthen the transferring ability of a detection model trained on source images.

Particularly, in the proposed detection network, we devise a progressive process to decompose the DIR and DSR with two disentangled layers. The goal of the first layer is to enhance the domain-invariant information in a middle-layer feature map. We utilize a domain classifier to ensure that DSR contains much more domain-specific information. And a mutual information (MI) loss is employed to enlarge the gap between DIR and DSR. Taking the sum of the feature map and DIR as the input, the second layer aims at obtaining the instance-invariant representations (IIR) with a regional proposal networks (RPN) \cite{ren2015faster,vu2019cascade}. Moreover, to enhance the disentanglement, we devise a training mechanism including three stages to optimize our model: (i) the stage of feature decomposition aiming at learning disentanglement, (ii) the stage of feature separation aiming at enlarging the gap between DIR and DSR, and (iii) the stage of feature reconstruction aiming at keeping the DIR and DSR contain all the content of the input. For each stage, we use different loss functions to optimize different components of our network, respectively. Experiments on three domain-shift scenes of DAOD demonstrate that our method is effective and achieves a new state-of-the-art performance.

The contributions of this paper are summarized as:

(1) Different from reducing the domain gap with distribution alignment, we propose to enhance the transferring detection ability via a bridge of disentangled instance-invariant representations.

(2) A progressive disentangled network is first proposed to successfully extract instance-invariant features. Meanwhile, a three-stage training mechanism is proposed to further enhance the disentangled ability.

(3) On three domain-shift scenes, i.e., Cityscapes \cite{cordts2016cityscapes} $\rightarrow$ FoggyCityscapes \cite{sakaridis2018semantic}, Pascal \cite{everingham2010pascal} $\rightarrow$ Watercolor \cite{inoue2018cross}, and Pascal $\rightarrow$ Clipart \cite{inoue2018cross}, our method is separately 2.3\%, 3.6\%, and 4.0\% higher than the baseline method \cite{saito2019strong}.
\vspace{-0.2in}
\section{Related Work}
\vspace{-0.1in}
\textbf{Domain Adaptive Object Detection.} Though most methods \cite{girshick2015fast,redmon2016you,hu2018relation,liu2016ssd} of object detection have achieved outstanding performance, their transferring abilities are limited for the task of DAOD. Recently, many methods \cite{kim2019diversify,saito2019strong,Kim2019SelfTrainingAA} have been proposed to solve the domain-shift problem in object detection. These methods mainly focus on feature-level or pixel-level alignment. For example, the method in \cite{chen2018domain} utilizes adversarial training \cite{ganin2014unsupervised} to align global feature distributions of the source and target domains, whereas the method in \cite{saito2019strong} aligns distributions of both global and local features. For pixel-level adaptation, the work \cite{kim2019diversify} devises a generative network to increase the diversity of the source domain, which is similar to data augmentation. However, as the alignment is conducted in holistic representations of images, it is not dedicated to the task of adaptive object detection, which focuses on the bridge of domains with instance-level characteristics. Therefore, in this paper, we focus on extracting instance-level features that are domain-invariant, which are helpful for improving the transferring ability of a detection method.

\begin{figure*}
\centering
\includegraphics[width=1.0\linewidth]{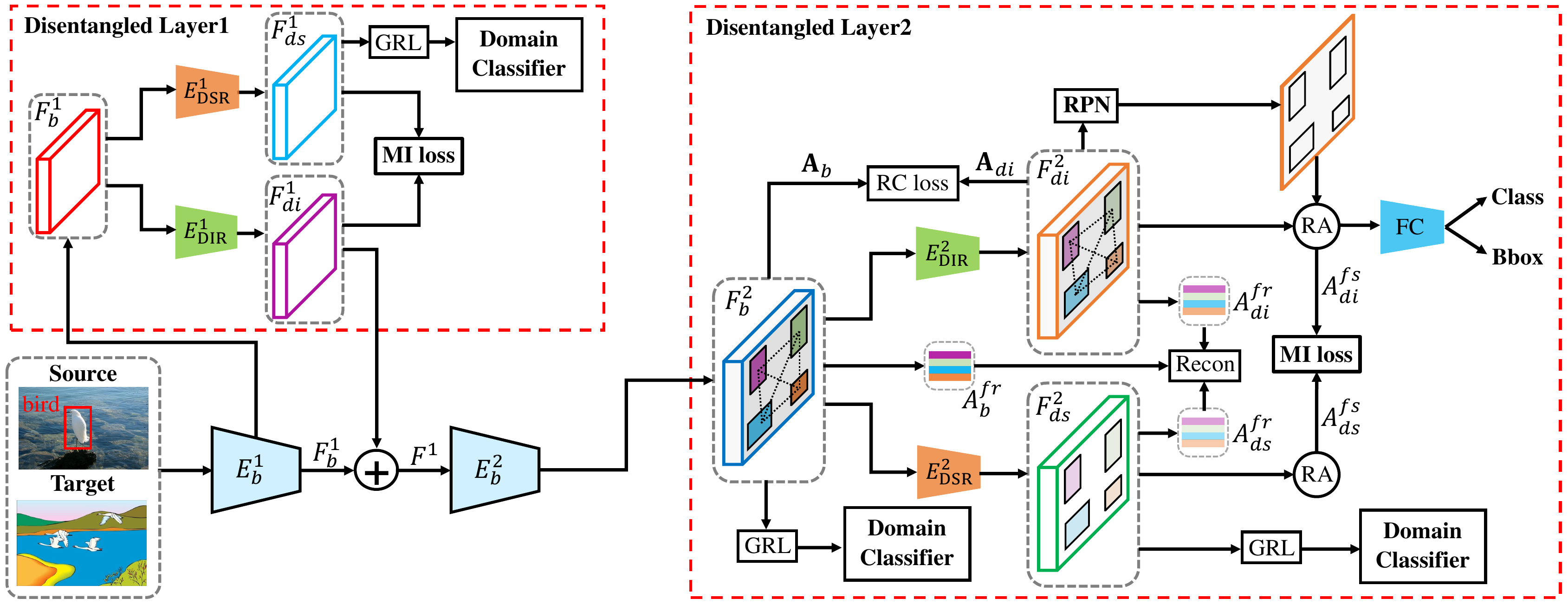}
\caption{Illustration of the proposed network of progressive disentanglement. `Recon' indicates the reconstruction loss. `GRL' is the gradient reverse layer. `RA' indicates the operation of RoI-Alignment. `RC loss' and `MI loss' separately denote the proposed relation-consistency loss and the mutual information loss. `$\oplus$' is the operation of element-wise sum. And the dot lines indicate the relations existed between the extracted proposals. There are two disentangled layers in the network. The purpose of the first layer is to enhance the domain-invariant information in a middle-layer feature map. And the goal of the second layer is to obtain the instance-invariant features. During training, in order to enhance the disentanglement, we devise a three-stage optimization mechanism with multiple loss functions. For each stage, we use different loss functions to optimize different components of the network.}
\label{fig2}
\vspace{-0.1in}
\end{figure*}

\textbf{Disentangled Learning.} The purpose of disentangled learning \cite{Jiang2018Graph,locatello2018challenging,bengio2013representation,peng2019domain} is to correctly uncover a set of independent factors that give rise to the current observation. Recently, disentangled learning has been well explored in tasks of few-shot learning \cite{ridgeway2018learning,scott2018adapted} and image translation \cite{lee2018diverse,huang2018multimodal}. Particularly, by decomposing the style of an image, the work \cite{lee2018diverse} proposed a disentangled method to make a diverse image-to-image translation. Liu et al. \cite{liu2018detach} proposed a model of cross-domain representation disentanglement. Based on generative adversarial networks, this method alleviated the impact of domain-shift and improved the classification performance on multiple datasets. As for adaptive object detection, on one hand, we should remove the domain-shift; on the other hand, it is important to transfer the detection ability via the bridge of the instance characteristics. Thus, it is not straightforward to apply the disentangled learning to the task of DAOD.

In this paper, we devise a new network of progressive disentanglement to decompose image representations into domain-specific and domain-invariant representations, and from which we extract the instance-invariant representations to bridge the detection ability between source and target domains. Experiments on three domain-shift scenes of DAOD demonstrate the effectiveness of our method.

\section{Instance-Invariant Adaptive Detection}

Suppose we have the access to an image $x^{s}$ including labels $y^{s}$ and bounding boxes $b^{s}$, which are drawn from a set of annotated source images $\{X_{s}, Y_{s}, B_{s}\}$. Here, $X_{s}$, $Y_{s}$, and $B_{s}$ separately indicate the set of images, labels, and bounding-box annotations, which are from the source domain. Meanwhile, we could also access to a target image $x^{t}$ drawn from a set of unlabeled target images $\{X_{t}\}$.

\subsection{The Network of Progressive Disentanglement}

As is shown in Fig. \ref{fig2}, we devise two disentangled layers to extract domain-invariant information progressively.

\textbf{The First Disentangled Layer.} The goal of this layer is to enhance the domain-invariant information in a middle-layer feature map. Concretely, given a source image $x^{s}$ and target image $x^{t}$, we first obtain a feature map $F_{b}^{1}$ that is the output of a middle-layer feature extractor ${E}_{b}^{1}$. Then, two different extractors are devised to disentangle the DIR and DSR from $F_{b}^{1}$. The processes are shown as follows:
\begin{equation}\label{DL1}
F_{di}^{1} = {E}_{\rm DIR}^{1}(F_{b}^{1}), F_{ds}^{1} = {E}_{\rm DSR}^{1}(F_{b}^{1}), F^{1} = F_{b}^{1} + F_{di}^{1}.
\end{equation}

Here, ${E}_{\rm DIR}^{1}$ and ${E}_{\rm DSR}^{1}$ separately indicate the DIR and DSR extractor. The size of $F_{di}^{1}$ and $F_{ds}^{1}$ is set to the same value as that of $F_{b}^{1}$. Then, we take the sum $F^{1}$ of $F_{di}^{1}$ and $F_{b}^{1}$ as the input of the second feature extractor ${E}_{b}^{2}$. Since $F_{di}^{1}$ contains more domain-invariant information, the sum operation could alleviate the impact of domain-shift on $F^{1}$.

\textbf{The Second Disentangled Layer.} The purpose of this layer is to obtain the instance-invariant features. Particularly, based on the output $F_{b}^{2}$ of the extractor ${E}_{b}^{2}$, we devise two extractors, i.e., ${E}_{\rm DIR}^{2}$ and ${E}_{\rm DSR}^{2}$, to disentangle the DIR and DSR from $F_{b}^{2}$. The processes are as follows:
\begin{equation}\label{DL2}
\begin{split}
& F_{b}^{2} = {E}_{b}^{2}(F^{1}), F_{di}^{2} = {E}_{\rm DIR}^{2}(F_{b}^{2}), F_{ds}^{2} = {E}_{\rm DSR}^{2}(F_{b}^{2}).
\end{split}
\end{equation}

Here, the size of $F_{di}^{2}$ and $F_{ds}^{2}$ is set to the same value as that of $F_{b}^{2}$. Next, the RPN is performed on $F_{di}^{2}$ to extract a set of instance-invariant proposals. Finally, for an image from the source domain, the detection loss is as follows:
\begin{equation}\label{class}
\begin{split}
& L_{D} = -\frac{1}{n_{s}}\sum_{j=1}^{n_{s}}L_{det}({D}(A_{j}),y_{j}^{s},b_{j}^{s}),
\end{split}
\end{equation}

\noindent where $n_{s}$ denotes the number of proposals. $A_{j}$ indicates the RoI-Alignment \cite{ren2015faster,he2017mask} result of the $j$-th proposal. ${D}$ includes the classification and regression network. $L_{det}$ is assumed to contain all the losses for the detection, e.g., classification and bounding-box regression loss.

\subsection{Training with the Three-stage Optimization}

\begin{figure}
\centering
\includegraphics[width=1.0\linewidth]{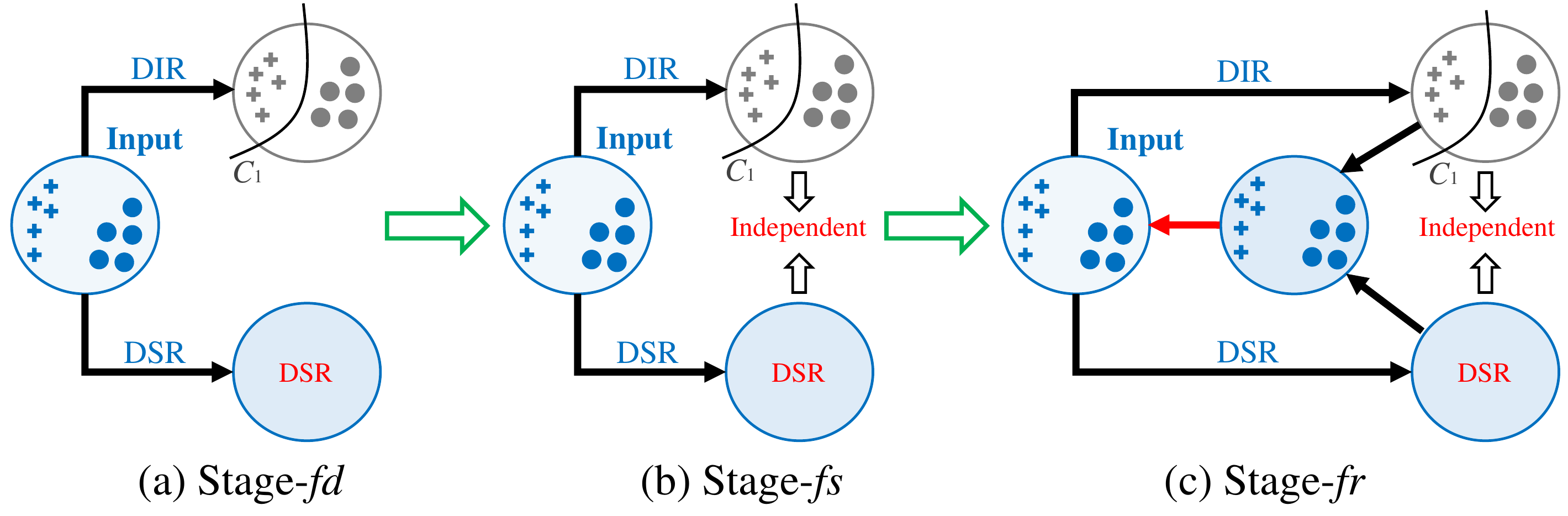}
\caption{Illustration of the three-stage training. Here, the red arrow denotes the operation of reconstruction. `Stage-$fd$' is the first stage aiming at learning disentanglement. `Stage-$fs$' is the second stage aiming at keeping the disentangled DIR and DSR independent. And `Stage-$fr$' is the third stage aiming at keeping the DIR and DSR could contain all the content of the input.}
\label{training}
\end{figure}

As is discussed in the section of Introduction, the goal of disentangled learning is to uncover a set of independent factors that give rise to the current observation \cite{do2019theory}. And these factors could contain all the information presents in the observation \cite{locatello2018challenging}. Therefore, we devise a three-stage training mechanism (see Fig. \ref{training}) to enhance the disentanglement.

\vspace{-0.8em}
\subsubsection{The Stage of Feature Decomposition}

The goal of the first stage is to ensure that our model not only learns the location and classification of the objects but also disentangles the image features. Based on $F_{di}^{2}$, we first utilize RPN to obtain a set of object proposals $O^{fd}$. To ensure that $F_{b}^{2}$ and $F_{di}^{2}$ have the same object contents in the same locations, based on the proposals $O^{fd}$, RoI-Alignment is performed on $F_{b}^{2}$ and $F_{di}^{2}$ to obtain $A_{b}^{fd}$ and $A_{di}^{fd}$, respectively. Next, we devise two networks ${D}_{b}$ and ${D}_{di}$ to perform the classification and bounding-box regression. Finally, for a source image, the detection loss is defined as:
\begin{equation}\label{fd}
\begin{split}
& L_{D}^{fd} = L_{D}^{b}(D_{b}(A_{b}^{fd})) + L_{D}^{di}(D_{di}(A_{di}^{fd})),
\end{split}
\end{equation}

\noindent where $L_{D}^{b}$ and $L_{D}^{di}$ indicate the detection loss.

By using the detection loss, $F_{b}^{2}$ and $F_{di}^{2}$ are ensured to contain the instance information. Besides, for our method, it is also important to keep the learned $F_{ds}^{1}$ and $F_{ds}^{2}$ contain more domain-specific information, which could ensure our model owns the ability of feature disentanglement. In this paper, we exploit the method of adversarial domain classification \cite{ganin2014unsupervised} to distinguish the source and target domains. Specifically, we employ four domain classifiers ${C}_{b}^{1}$, ${C}_{ds}^{1}$, ${C}_{b}^{2}$, and ${C}_{ds}^{2}$ in our model, which separately take $F_{b}^{1}$, $F_{ds}^{1}$, $F_{b}^{2}$, and $F_{ds}^{2}$ as the input and output a domain label $l_{d}$ that indicates the source or target domain: $l_{d}$ is 0 for the source domain and 1 for the target domain.

Besides, for domain classifiers, during training, we employ Focal Loss ($FL$) \cite{lin2017focal,saito2019strong} to impose bigger weights on the hard-to-classify examples (i.e., the examples near the classification boundary) than on the easy ones (i.e., the examples far from the classification boundary).
\begin{equation}\label{FL}
\begin{split}
& {FL}(p) = -g(p)log(p),\quad g(p) = \alpha(1-p)^{\gamma},
\end{split}
\end{equation}

\noindent where $\gamma$ controls the weight on the hard-to-classify examples. $p \in [0, 1]$ is the model's estimated probability for the output domain label $l_{d}$. Finally, the loss of the first training stage is denoted as follows:
\begin{equation}\label{loss1}
\begin{split}
& L_{s}^{fd} = L_{D}^{fd} + {FL}_{s}({C}_{b}^{1}(F_{b}^{1})) + {FL}_{s}({C}_{ds}^{1}(F_{ds}^{1}))\\
& \qquad\quad + {FL}_{s}({C}_{b}^{2}(F_{b}^{2})) + {FL}_{s}({C}_{ds}^{2}(F_{ds}^{2})), \\
& L_{t}^{fd} = {FL}_{t}({C}_{b}^{1}(F_{b}^{1})) + {FL}_{t}({C}_{ds}^{1}(F_{ds}^{1})) \\
& \qquad\quad + {FL}_{t}({C}_{b}^{2}(F_{b}^{2})) + {FL}_{t}({C}_{ds}^{2}(F_{ds}^{2})),
\end{split}
\end{equation}

\noindent where $L_{s}^{fd}$ and $L_{t}^{fd}$ are the objective functions of the source and target domains. ${FL}_{s}$ and ${FL}_{t}$ indicate the domain losses. The overall loss $L^{fd}$ is the sum of $L_{s}^{fd}$ and $L_{t}^{fd}$.

With the help of the detection loss $L_{D}^{fd}$ and domain loss $FL$, the disentangled DIR and DSR contain instance and domain-specific information, respectively. Next, we will perform the second training stage to keep the disentangled DIR and DSR independent.

\vspace{-0.8em}
\subsubsection{The Stage of Feature Separation}

In this stage, we first fix the extractor ${E}_{b}^{1}$ and ${E}_{b}^{2}$ of the model trained on the first stage. Then, we employ the model to extract $F_{b}^{1}$, $F_{di}^{1}$, $F_{ds}^{1}$ (Eq. \eqref{DL1}), $F_{b}^{2}$, $F_{di}^{2}$, and $F_{ds}^{2}$ (Eq. \eqref{DL2}). RPN is performed on $F_{di}^{2}$ to obtain the proposals $O^{fs}$.

\textbf{Mutual Information Minimization.} In order to enlarge the gap between the DIR and DSR, we minimize the MI loss between $A_{di}^{fs}$ and $A_{ds}^{fs}$, as well as between $F_{di}^{1}$ and $F_{ds}^{1}$, where $A_{di}^{fs}$ and $A_{ds}^{fs}$ indicate the RoI-Alignment results of $F_{di}^{2}$ and $F_{ds}^{2}$ based on $O^{fs}$. The process of MI is:
\begin{equation}\label{MI}
\begin{split}
& I(X;Z) = \int_{X \times Z}\log\frac{d\mathbb{P}_{XZ}}{d\mathbb{P}_{X}\otimes \mathbb{P}_{Z}}d\mathbb{P}_{XZ},
\end{split}
\end{equation}

\noindent where $\mathbb{P}_{XZ}$ indicates the joint probability distribution of ($A_{di}^{fs}$, $A_{ds}^{fs}$) or ($F_{di}^{1}$, $F_{ds}^{1}$). $\mathbb{P}_{X} = \int_{Z}d\mathbb{P}_{XZ}$ and $\mathbb{P}_{Z} = \int_{X}d\mathbb{P}_{XZ}$ are the marginal distributions. Obviously, by minimizing the MI loss, we could impose independent constraints on the tuples ($A_{di}^{fs}$, $A_{ds}^{fs}$) and ($F_{di}^{1}$, $F_{ds}^{1}$). Besides, since $F_{ds}^{1}$ and $F_{ds}^{2}$ contain more domain-specific information, MI loss could promote $F_{di}^{1}$ and $F_{di}^{2}$ to contain more domain-invariant information, which can help strengthen the ability of disentanglement. In this paper, we adopt Mutual Information Neural Estimator (MINE) \cite{belghazi2018mine} to compute the MI loss. Concretely, based on Monte-Carlo integration \cite{peng2019domain}, MINE could be computed as follows:
\begin{equation}\label{MI2}
\begin{split}
& I(X,Z) = \frac{1}{n}\sum_{j=1}^{n}{T}(x,z,\theta)-\log(\frac{1}{n}\sum_{j=1}^{n}e^{{T}(x,z',\theta)}),
\end{split}
\end{equation}

\noindent where $(x,z)$ is sampled from the joint distribution and $z'$ is sampled from the marginal distribution. Here, we devise a neural network to perform the Monte-Carlo integration.

It is worth noting that, for the second disentangled layer, we use the RoI-Alignment results $A_{di}^{fs}$ and $A_{ds}^{fs}$, instead of the feature map $F_{di}^{2}$ and $F_{ds}^{2}$, to compute MI loss, which could not only reduce the computational costs but also ensure our model pays more attention to regions of objects.

\begin{figure}
\centering
\includegraphics[width=1.0\linewidth]{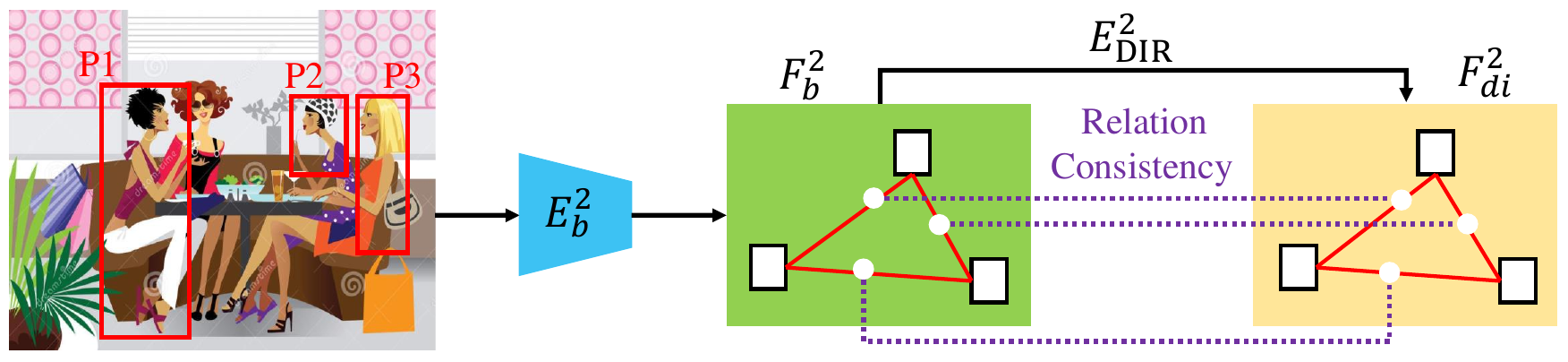}
\caption{Illustration of relation-consistency Loss. `P' indicates the `Person' class. The goal of the loss is to ensure the relations (the red solid lines) between object proposals in $F_{b}^{2}$ and the relations between object proposals in $F_{di}^{2}$ are consistent. The purple dot lines denote the consistency between the two red lines.}
\label{fig3}
\end{figure}

\textbf{Relation-consistency Loss.} To further improve the disentanglement, we devise a relation-consistency loss (Fig. \ref{fig3}). Specifically, since $F_{di}^{2}$ and $F_{b}^{2}$ have the same object contents in the same locations, based on the proposals $O^{fs}$, $A_{di}^{fs}$ and $A_{b}^{fs}$ should keep similar semantic relations.

Concretely, we first obtain the average-pooling results $P_{di}^{fs} \in \mathbb{R}^{k \times m}$ and $P_{b}^{fs} \in \mathbb{R}^{k \times m}$ of $A_{di}^{fs}$ and $A_{b}^{fs}$, where $k$ and $m$ indicate the numbers of proposals and channels. Then we separately construct a graph $G_{di}=\{\mathcal{V}_{di}, \mathcal{E}_{di}\}$ and $G_{b}=\{\mathcal{V}_{b}, \mathcal{E}_{b}\}$. Here, we take $P_{di}^{fs}$ and $P_{b}^{fs}$ as the nodes $\mathcal{V}_{di}$ and $\mathcal{V}_{b}$, respectively. $\mathcal{E}_{di}$ and $\mathcal{E}_{b}$ are used to indicate the edges (relations) between proposals. Next, we define two adjacency matrix for two undirected graphs, i.e., $\textbf{A}_{b} = softmax_{r}((P_{b}^{fs})(P_{b}^{fs})^{\rm T})$ and $\textbf{A}_{di} = softmax_{r}((P_{di}^{fs})(P_{di}^{fs})^{\rm T})$. And $softmax_{r}$ indicates we make $softmax$ operation across the row directions. The relation-consistency loss is computed as:
\begin{equation}\label{relation}
\begin{split}
& L_{rel} = ||\textbf{A}_{di}-\textbf{A}_{b}||_{2}^{2}.
\end{split}
\end{equation}

Note that the computation of the relation-consistency loss does not need any parameters. Finally, the loss of the second training stage is denoted as follows:
\begin{equation}\label{second}
\begin{split}
& L_{s}^{fs} = L_{D}^{di}(D_{di}(A_{di}^{fs})) + {FL}_{s}({C}_{ds}^{2}(F_{ds}^{2})) + I_{s}^{2} \\
& \qquad\quad + L_{rel}^{s} + {FL}_{s}({C}_{ds}^{1}(F_{ds}^{1})) + I_{s}^{1}, \\
& L_{t}^{fs} = {FL}_{t}({C}_{ds}^{2}(F_{ds}^{2})) + I_{t}^{2} + L_{rel}^{t} \\
& \qquad\quad + {FL}_{t}({C}_{ds}^{1}(F_{ds}^{1})) + I_{t}^{1},
\end{split}
\end{equation}

\noindent where $L_{D}^{di}$ is the detection loss based on $A_{di}^{fs}$. $L_{s}^{fs}$ and $L_{t}^{fs}$ are the training objectives of the source and target domain, respectively. $I^{1}$ and $I^{2}$ indicate MI loss computed on the first and second disentangled layer, respectively. The overall loss $L^{fs}$ is the sum of $L_{s}^{fs}$ and $L_{t}^{fs}$. After this stage, the gap between DIR and DSR could be enlarged. Next, we will perform the third training stage aiming at keeping the disentangled DIR and DSR contain all the content of the input used for disentanglement.

\begin{algorithm}[t]
  \caption{Instance-Invariant Adaptive Object Detection} \label{alg_DAOD}
  \begin{small}
      \hspace*{0.02in}{\bf Input}: source images $\{X_{s}, Y_{s}, B_{s}\}$; target images $\{X_{t}\}$; feature extractors ${E}_{b}^{1}$ and ${E}_{b}^{2}$; disentangled extractors ${E}_{\rm DIR}^{1}$, ${E}_{\rm DSR}^{1}$, ${E}_{\rm DIR}^{2}$, and ${E}_{\rm DSR}^{2}$; detection networks ${D}_{b}$ and ${D}_{di}$; domain classifiers ${C}_{b}^{1}$, ${C}_{ds}^{1}$, ${C}_{b}^{2}$, and ${C}_{ds}^{2}$; MI estimators ${T}^{1}$ and ${T}^{2}$; reconstruction network ${R}$. \\
      \hspace*{0.02in}{\bf Output}: trained ${\hat{E}}_{b}^{1}$, ${\hat{E}}_{\rm DIR}^{1}$, ${\hat{E}}_{\rm DSR}^{1}$, ${\hat{E}}_{b}^{2}$, ${\hat{E}}_{\rm DIR}^{2}$, ${\hat{E}}_{\rm DSR}^{2}$, and detection networks $\hat{D}_{b}$, $\hat{D}_{di}$.
      \begin{algorithmic}[1]
          \WHILE {not converged}
          \STATE  Sample a mini-batch from $\{X_{s}, Y_{s}, B_{s}\}$ and $\{X_{t}\}$;
          \STATE \textbf{Feature Decomposition:}
          \STATE Update ${E}_{b}^{1}$, ${E}_{\rm DIR}^{1}$, ${E}_{b}^{2}$, ${E}_{\rm DIR}^{2}$, ${D}_{b}$, ${D}_{di}$ by Eq. \eqref{fd};
          \STATE Update ${E}_{b}^{1}$, ${E}_{\rm DSR}^{1}$, ${C}_{b}^{1}$, ${C}_{ds}^{1}$, ${E}_{b}^{2}$, ${E}_{\rm DSR}^{2}$, ${C}_{b}^{2}$, and ${C}_{ds}^{2}$ by Eq. \eqref{FL};
          \STATE \textbf{Feature Separation:}
          \STATE Update ${E}_{\rm DIR}^{1}$, ${E}_{\rm DIR}^{2}$, ${D}_{di}$ by $L_{D}^{di}$ in Eq. \eqref{second};
          \STATE Update ${E}_{\rm DSR}^{1}$, ${C}_{ds}^{1}$, ${E}_{\rm DSR}^{2}$, ${C}_{ds}^{2}$ by ${FL}$ in Eq. \eqref{second};
          \STATE Calculate the MI loss between $F_{di}^{1}$ and $F_{ds}^{1}$ with ${T^{1}}$, and between $A_{di}^{fs}$ and $A_{ds}^{fs}$ with ${T^{2}}$;
          \STATE Update ${E}_{\rm DIR}^{1}$, ${E}_{\rm DSR}^{1}$, ${T^{1}}$, ${E}_{\rm DIR}^{2}$, ${E}_{\rm DSR}^{2}$, ${T^{2}}$ by Eq. \eqref{MI2};
          \STATE Update ${E}_{\rm DIR}^{2}$ by Eq. \eqref{relation};
          \STATE \textbf{Feature Reconstruction:}
          \STATE Reconstruct RoI-Alignment result $A_{b}^{fr}$ by ($A_{di}^{fr}$, $A_{ds}^{fr}$);
          \STATE Update ${E}_{\rm DIR}^{2}$, ${E}_{\rm DSR}^{2}$, ${R}$ by Eq. \eqref{third};
          \ENDWHILE
          \RETURN $\hat{E}_{b}^{1}={E}_{b}^{1}; \hat{E}_{\rm DIR}^{1}={E}_{\rm DIR}^{1}; \hat{E}_{\rm DSR}^{1}={E}_{\rm DSR}^{1}; \hat{E}_{b}^{2}={E}_{b}^{2}; \hat{E}_{\rm DIR}^{2}={E}_{\rm DIR}^{2}; \hat{E}_{\rm DSR}^{2}={E}_{\rm DSR}^{2}; \hat{D}_{b}={D}_{b}; \hat{D}_{di}={D}_{di}$.
      \end{algorithmic}
    \end{small}
\end{algorithm}

\vspace{-0.8em}
\subsubsection{The Stage of Feature Reconstruction}

We employ a reconstruction loss to attain the purpose of this training stage. Concretely, we first use the model trained on the second stage to extract $F_{b}^{2}$, $F_{di}^{2}$, and $F_{ds}^{2}$ (Eq. \eqref{DL2}). Then, RPN is performed on $F_{di}^{2}$ to extract proposals $O^{fr}$. The reconstruction loss is computed as follows:
\begin{equation}\label{third}
\begin{split}
&A_{r}^{fr} = {R}(\langle A_{di}^{fr}, A_{ds}^{fr} \rangle), \quad L_{recon} = ||A_{r}^{fr} - A_{b}^{fr}||_{2}^{2},
\end{split}
\end{equation}

\noindent where $A_{di}^{fr}$, $A_{ds}^{fr}$, and $A_{b}^{fr}$ are the RoI-Alignment results of $F_{di}^{2}$, $F_{ds}^{2}$, and $F_{b}^{2}$ based on the proposals $O^{fr}$. ${R}$ is the reconstruction network. $\langle a, b \rangle$ indicates the concatenation of $a$ and $b$. Here, in order to make the model pay more attention to instance content, the reconstruction loss is only computed on the regions of the proposals. Besides, since the output of the first disentangled layer includes the entire $F_{b}^{1}$, to reduce the computational costs, we do not calculate the reconstruction loss on the first layer.

\begin{table*}
\small
\begin{center}
\begin{tabular}{l|c|cccccccc|c}
\toprule[1.5pt]
Method  & backbone & person & rider & car & truck & bus & train & motorcycle & bicycle & mAP \\
\hline
Source Only & VGG16 & 24.7 & 31.9 & 33.1 & 11.0 & 26.4 & 9.2 & 18.0 & 27.9 & 22.8 \\
\hline
DAF \cite{chen2018domain} & VGG16 & 25.0 & 31.0 & 40.5 & 22.1 & 35.3 & 20.2 & 20.0 & 27.1 & 27.6 \\ \hline
DT \cite{inoue2018cross} & VGG16 & 25.4 & 39.3 & 42.4 & 24.9 & 40.4 & 23.1 & 25.9 & 30.4 & 31.5 \\ \hline
SC-DA(Type3) \cite{zhu2019adapting} & VGG16 & 33.5 & 38.0 & 48.5 & 26.5 & 39.0 & 23.3 & 28.0 & 33.6 & 33.8 \\ \hline
DMRL \cite{kim2019diversify} & VGG16 & 30.8 & 40.5 & 44.3 & 27.2 & 38.4 & 34.5 & 28.4 & 32.2 & 34.6 \\ \hline
MTOR \cite{cai2019exploring} & ResNet50 & 30.6 & 41.4 & 44.0 & 21.9 & 38.6 & {\bf 40.6} & 28.3 & 35.6 & 35.1 \\ \hline
MLDA \cite{xie2019multi} & VGG16 & 33.2 & 44.2 & 44.8 & 28.2 & 41.8 & 28.7 & {\bf 30.5} & 36.5 & 36.0 \\ \hline
FSDA \cite{wang2019few} & VGG16 & 29.1 & 39.7 & 42.9 & 20.8 & 37.4 & 24.1 & 26.5 & 29.9 & 31.3 \\ \hline
MAF \cite{he2019multi} & VGG16 & 28.2 & 39.5 & 43.9 & 23.8 & 39.9 & 33.3 & 29.2 & 33.9 & 34.0 \\ \hline
RLDA \cite{khodabandeh2019robust} & IncepV2 \cite{szegedy2016rethinking} & {\bf 35.10} & 42.15 & 49.17 & 30.07 & 45.25 & 26.97 & 26.85 & 36.03 & 36.45 \\ \hline \hline
SW (B) \cite{saito2019strong} & VGG16 & 29.9 & 42.3 & 43.5 & 24.5 & 36.2 & 32.6 & 30.0 & 35.3 & 34.3 \\
Ours & VGG16 & 33.12 & 43.41 & {\bf 49.63} & 21.98 & 45.75 & 32.04 & 29.59 & {\bf 37.08} & 36.57 \\
Ours & ResNet101 & 32.82 & {\bf 44.37} & 49.57 & {\bf 33.02} & {\bf 46.10} & 37.97 & 29.90 & 35.26 & {\bf 38.63} \\
\bottomrule[1.5pt]
\end{tabular}
\caption{Results (\%) on adaptation from Cityscapes to FoggyCityscapes. `B' indicates the baseline method. `Source Only' indicates the model is only trained based on the data from the source domain and does not use the target data.}\label{cityscape}
\end{center}
\end{table*}

In this paper, our model is trained in an end-to-end way. The detailed training procedures are presented in Algorithm \ref{alg_DAOD}. During each training stage, the parameters that do not appear in the current stage are considered to be fixed.

\section{Experiments}

We evaluate our approach on three domain-shift scenes, i.e., Cityscapes \cite{cordts2016cityscapes} $\rightarrow$ FoggyCityscapes \cite{sakaridis2018semantic}, Pascal VOC \cite{everingham2010pascal} $\rightarrow$ Watercolor \cite{inoue2018cross}, and Pascal VOC $\rightarrow$ Clipart \cite{inoue2018cross}. 

\subsection{Dataset and Implementation Details}

\textbf{Dataset.} For Cityscapes $\rightarrow$ FoggyCityscapes, we use Cityscapes as the source domain. FoggyCityscapes is used as the target domain, which is rendered from Cityscapes and simulates the change of weather condition. Both of them contain 2,975 images in the training set and 500 images in the validation set. And this adaptation scene involves 8 categories. We utilize the training set during training and evaluate on the validation set.

For Pascal $\rightarrow$ Watercolor and Pascal $\rightarrow$ Clipart, Pascal VOC dataset is used as the real source domain. The images of this dataset include rich bounding box annotations. And the number of object classes is 20. Following a prevalent setting \cite{kim2019diversify,saito2019strong}, we use Pascal VOC 2007 and 2012 training and validation set for training, which results in about 15K images. Watercolor and Clipart datasets are taken as the target domain. Watercolor contains 6 categories in common with VOC and 2k images in total. Clipart contains 1k images in total, which has the same 20 categories as VOC. For these two target datasets, the splits of training and test set are the same as the work \cite{saito2019strong}.

\textbf{Implementation Details.} Our method is based on Faster-RCNN \cite{ren2015faster} with RoI-Alignment \cite{he2017mask}. For Focal Loss (Eq. \eqref{FL}), $\alpha$ and $\gamma$ are set to 1.0 and 2.0. Besides, we separately employ a network including three convolutional layers as the disentangled extractors $E_{\rm DIR}^{1}$, $E_{\rm DSR}^{1}$, $E_{\rm DIR}^{2}$, and $E_{\rm DSR}^{2}$. For the domain classifiers $C_{b}^{1}$, $C_{ds}^{1}$, $C_{b}^{2}$, and $C_{ds}^{2}$, we respectively employ a network which includes three fully-connected layers. Meanwhile, for the MI estimators $T^{1}$ and $T^{2}$, we separately utilize a network consisting of three fully-connected layers. Finally, one convolutional layer is used as the reconstruction network ${R}$. During training, we employ the SGD optimizer with momentum \cite{Sutskever2013On}. We first train the model with a learning rate of 0.001 for 50K iterations, then with a learning rate of 0.0001 for 30K more iterations. In the test, we use mean average precisions (mAP) as the evaluation metric.

\begin{figure*}[ht]
\begin{center}
  \subfigure[\footnotesize Raw image in Cityscapes]{
  \begin{minipage}[t]{0.18\linewidth}
    \includegraphics[width=1.30in]{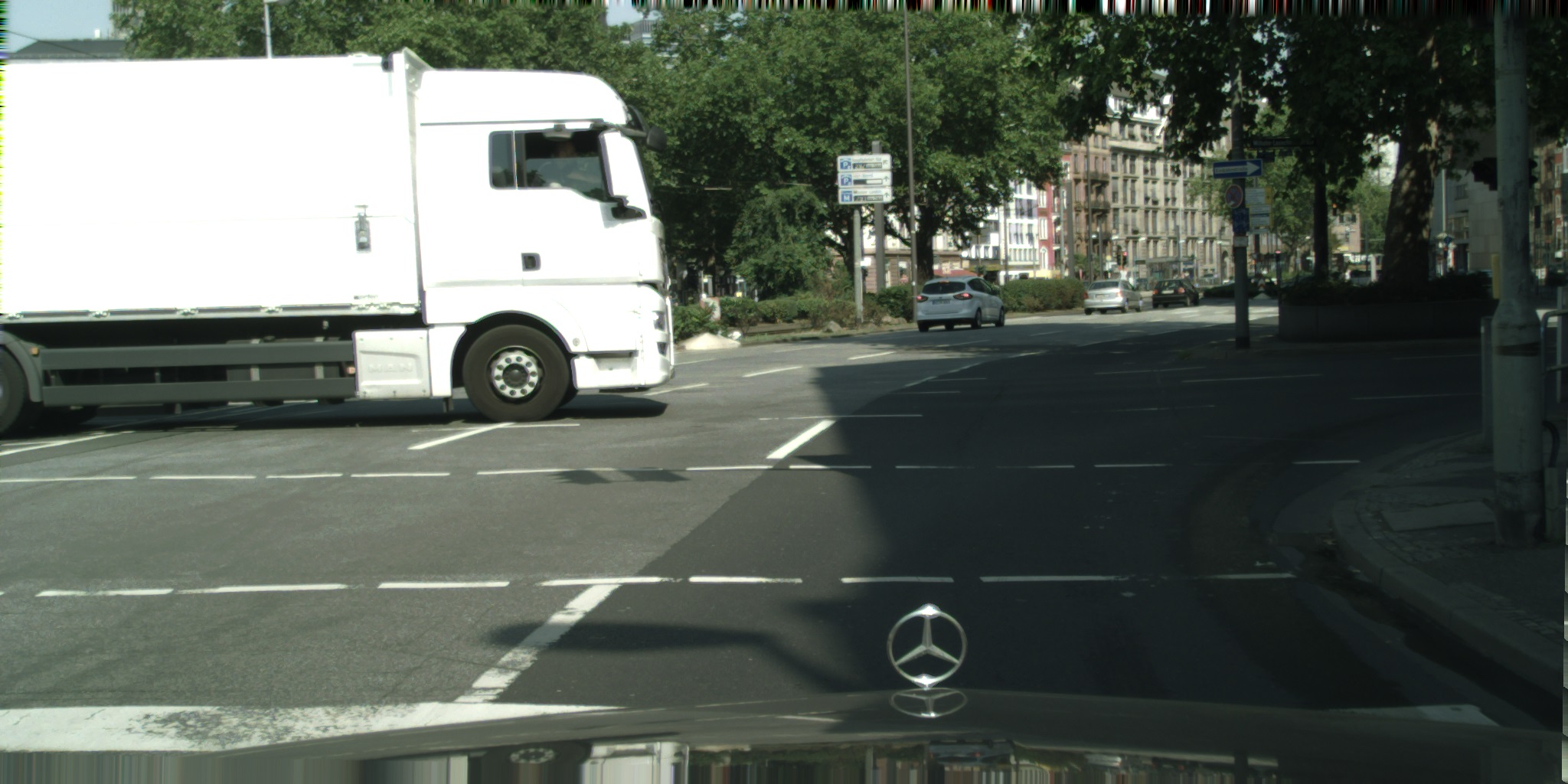}\\ \vspace{-0.1in}
    \includegraphics[width=1.30in]{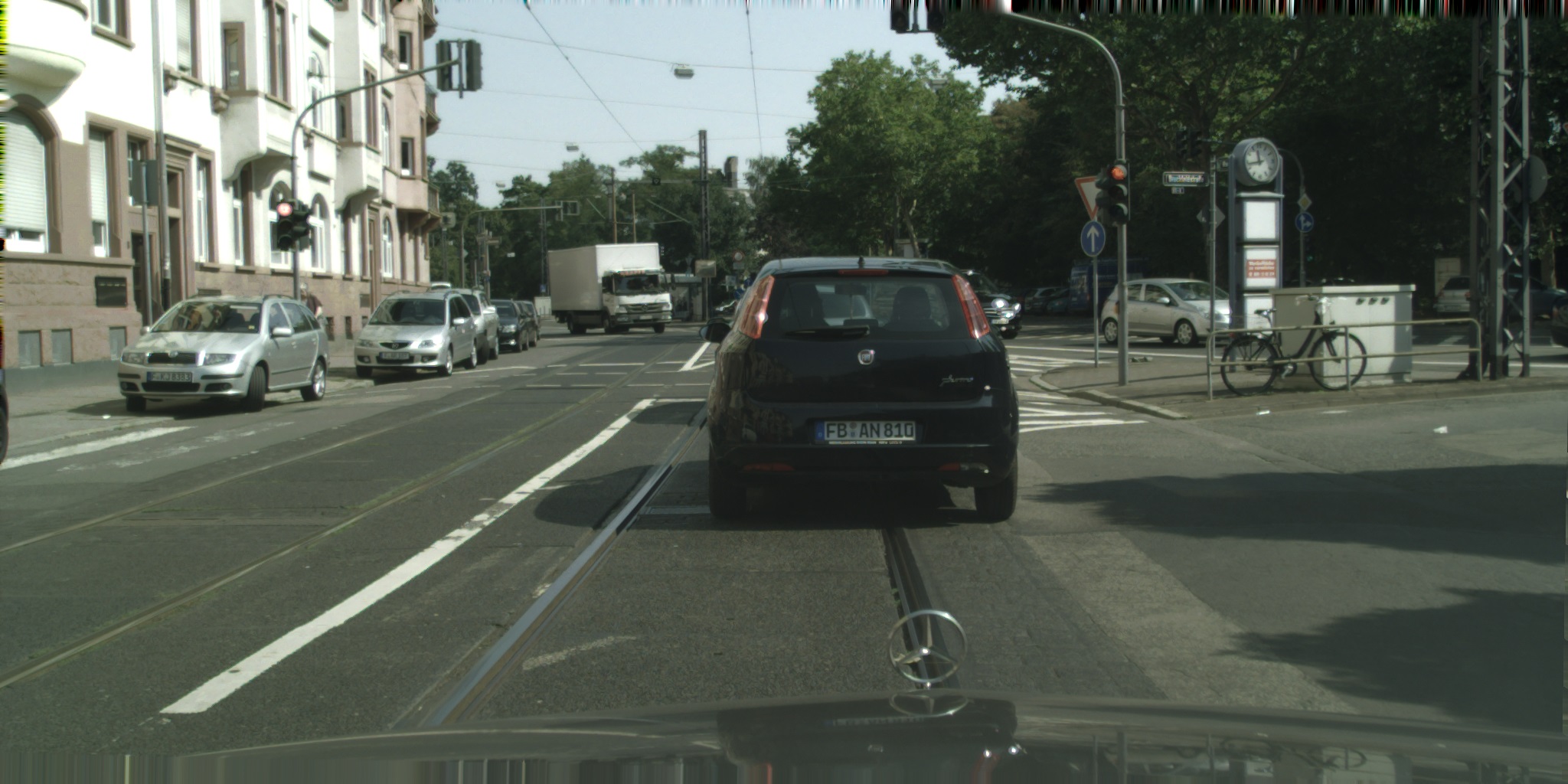}
  \end{minipage}
  }
  \hspace{0.02cm}
  \subfigure[\footnotesize GT]{
  \begin{minipage}[t]{0.18\linewidth}
    \includegraphics[width=1.30in]{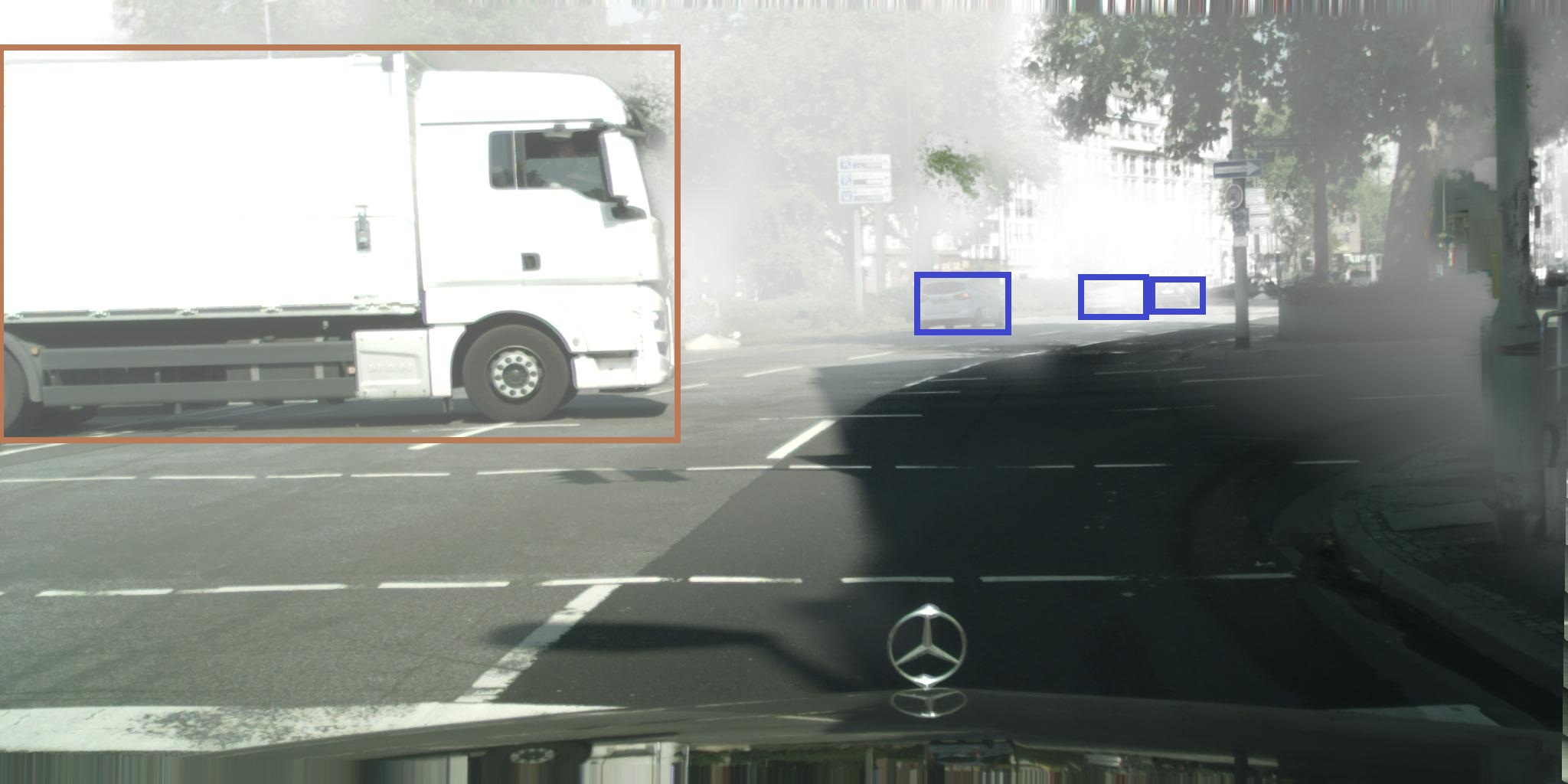}\\ \vspace{-0.1in}
    \includegraphics[width=1.30in]{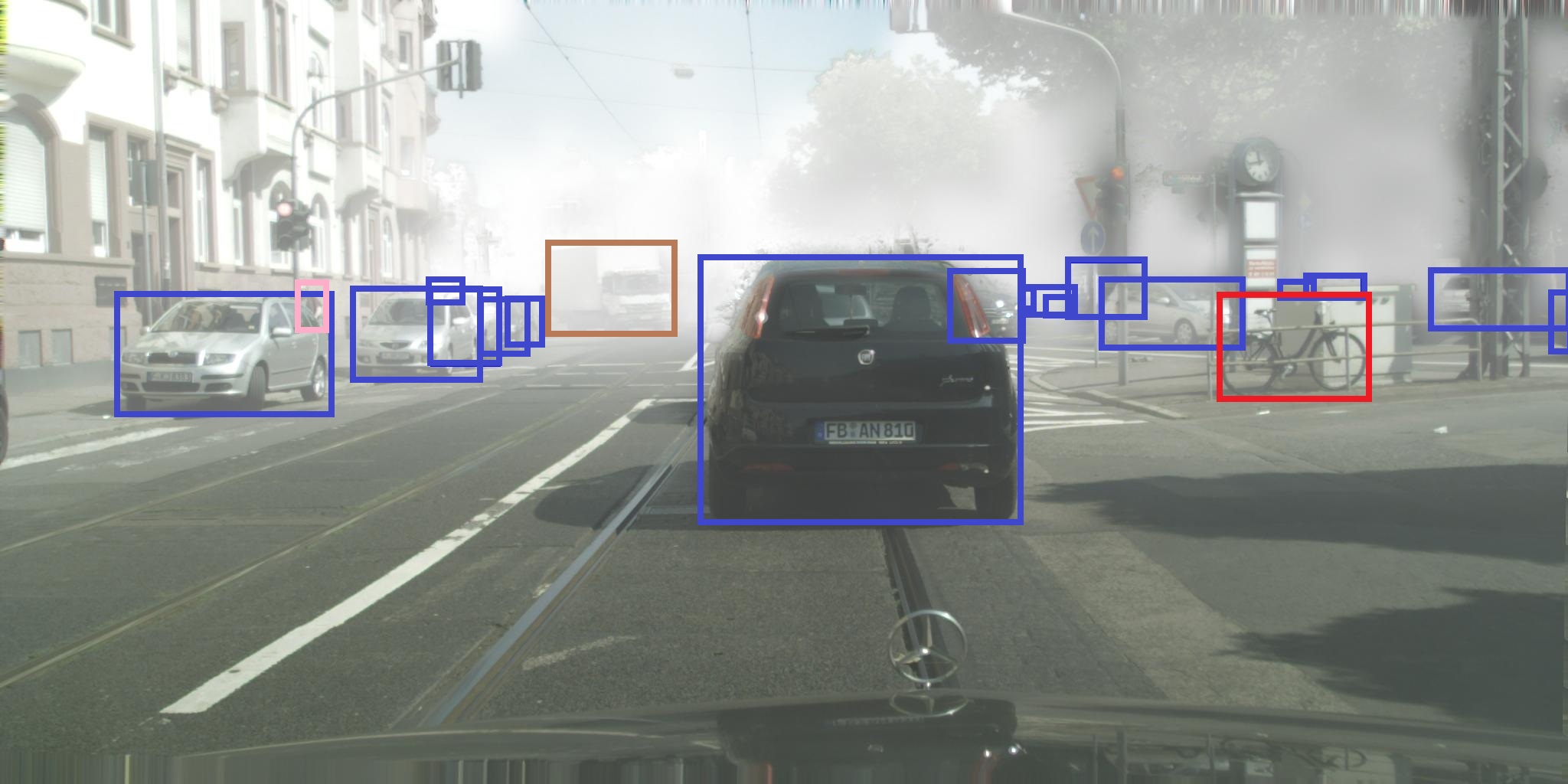}
  \end{minipage}
  }
  \hspace{0.02cm}
  \subfigure[\footnotesize SW baseline]{
  \begin{minipage}[t]{0.18\linewidth}
    \includegraphics[width=1.30in]{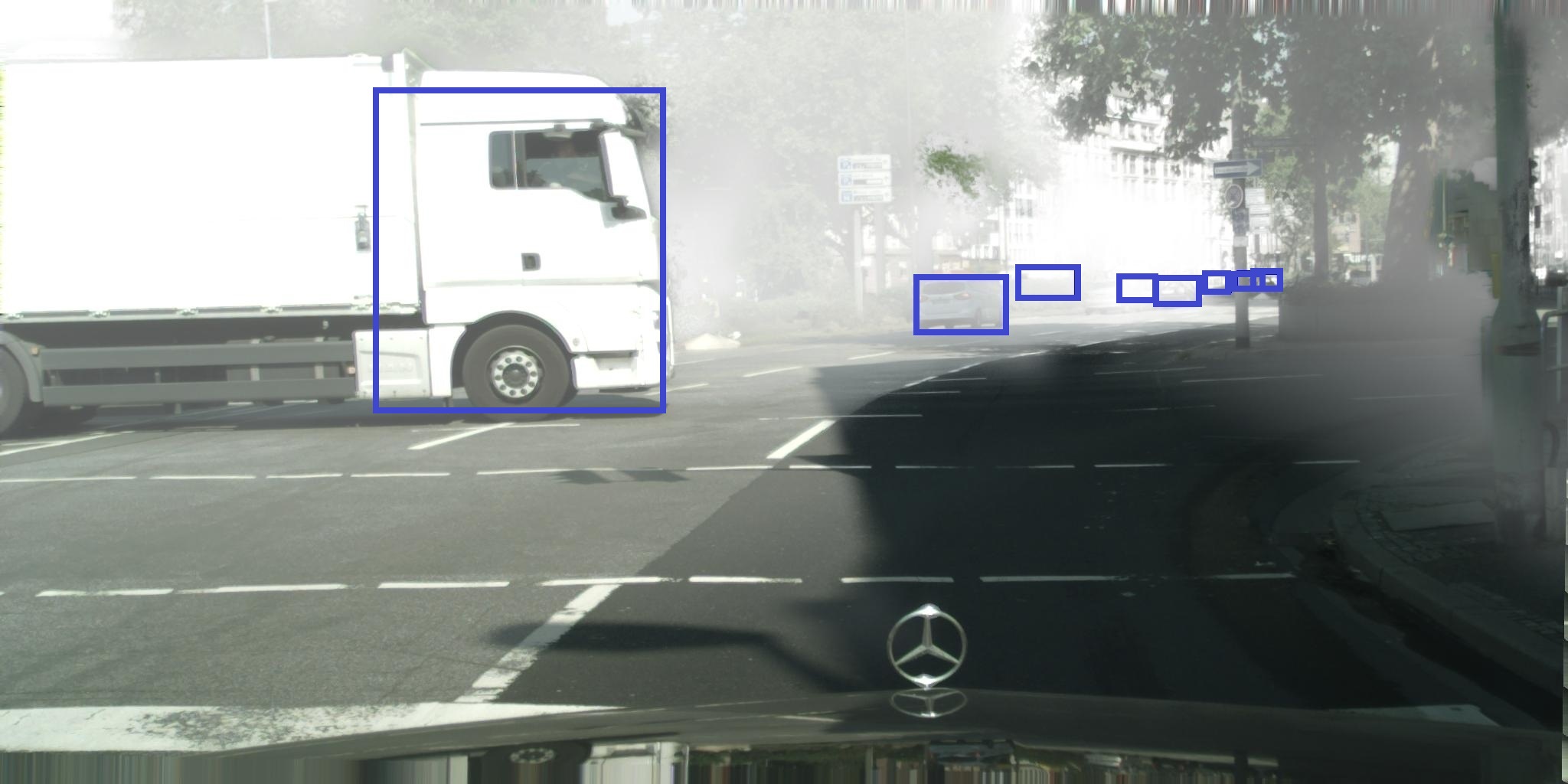}\\ \vspace{-0.1in}
    \includegraphics[width=1.30in]{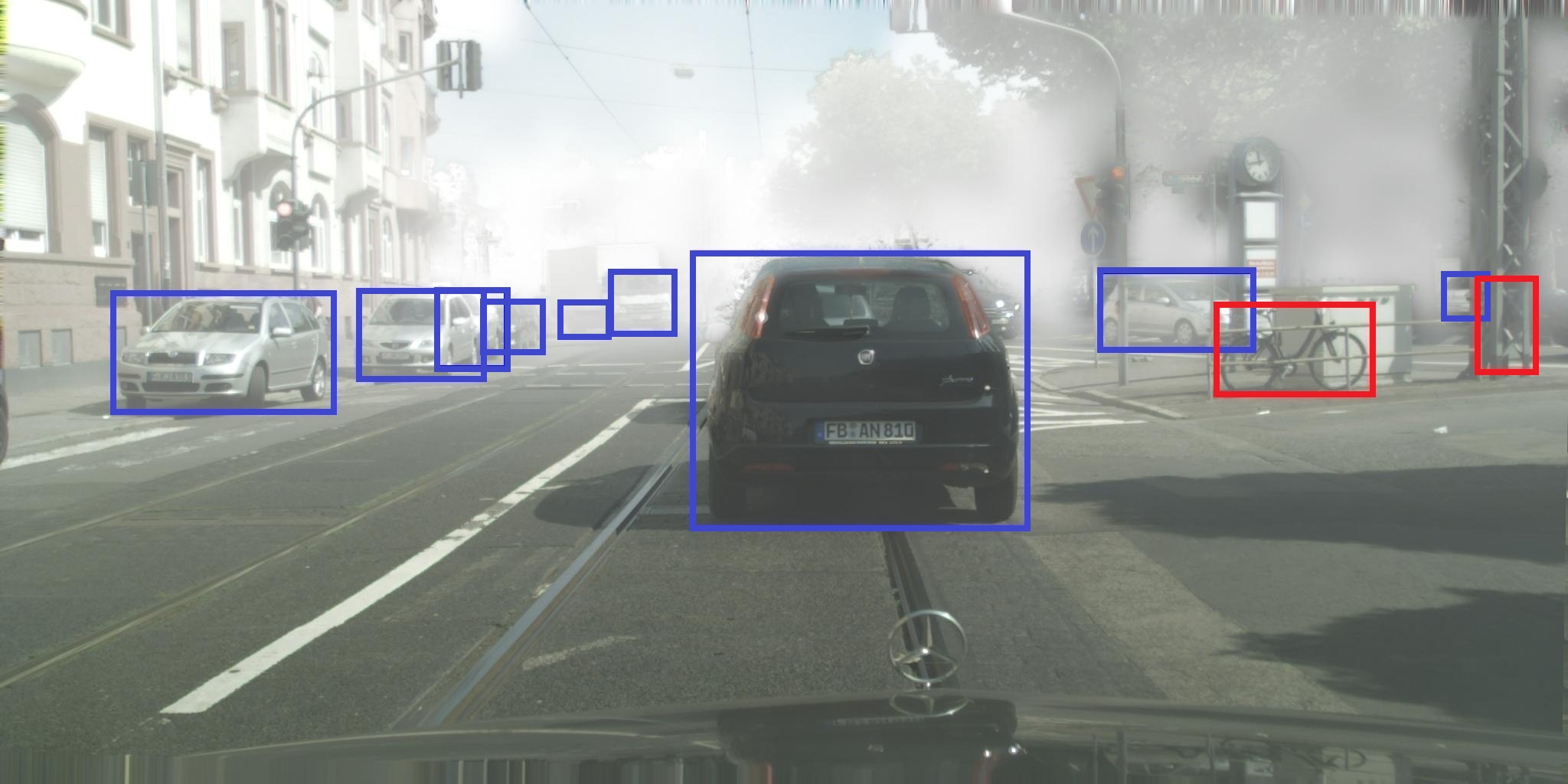}
  \end{minipage}
  }
  \hspace{0.02cm}
  \subfigure[\footnotesize One Disentangled layer]{
  \begin{minipage}[t]{0.18\linewidth}
    \includegraphics[width=1.30in]{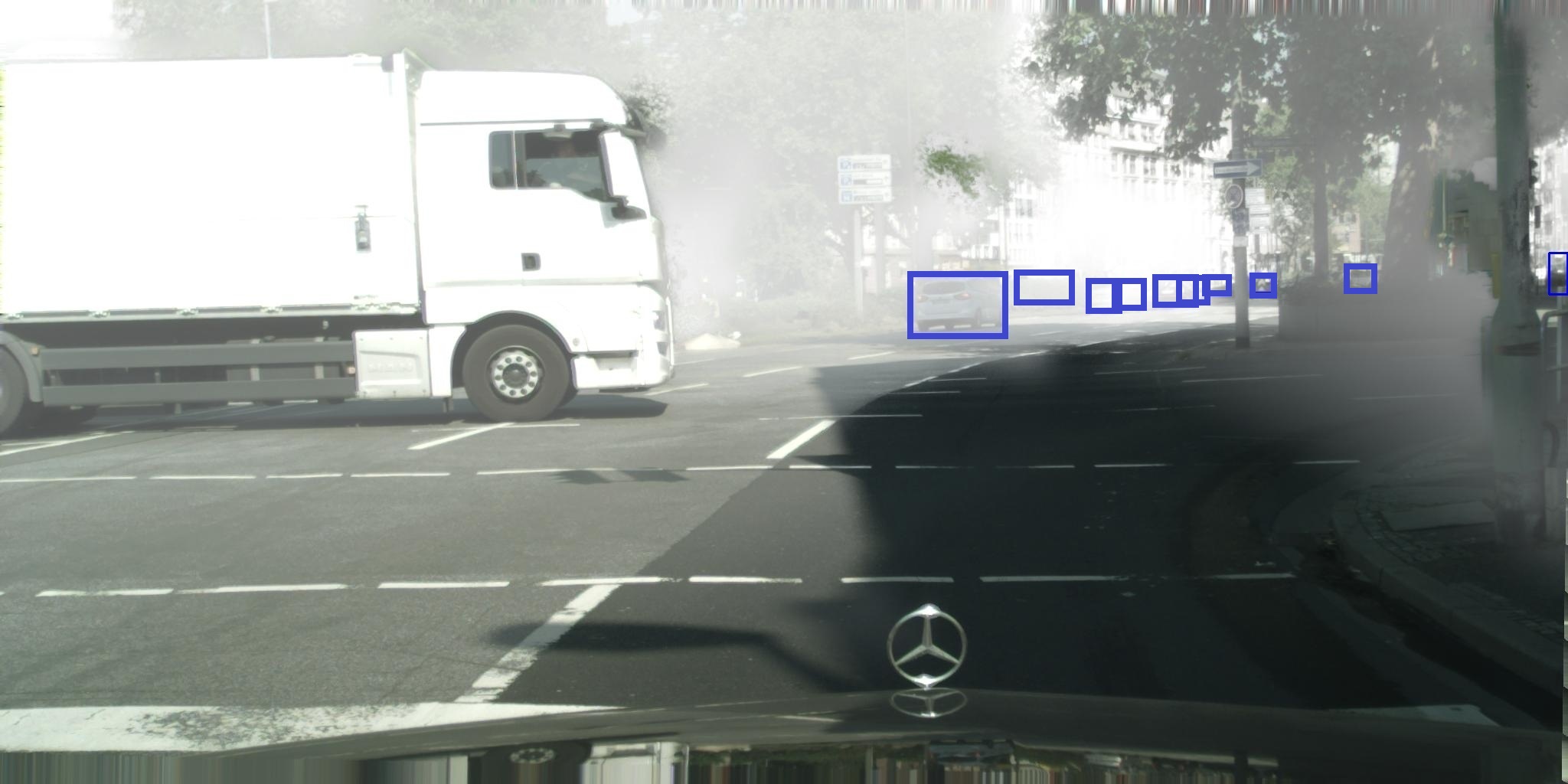}\\ \vspace{-0.1in}
    \includegraphics[width=1.30in]{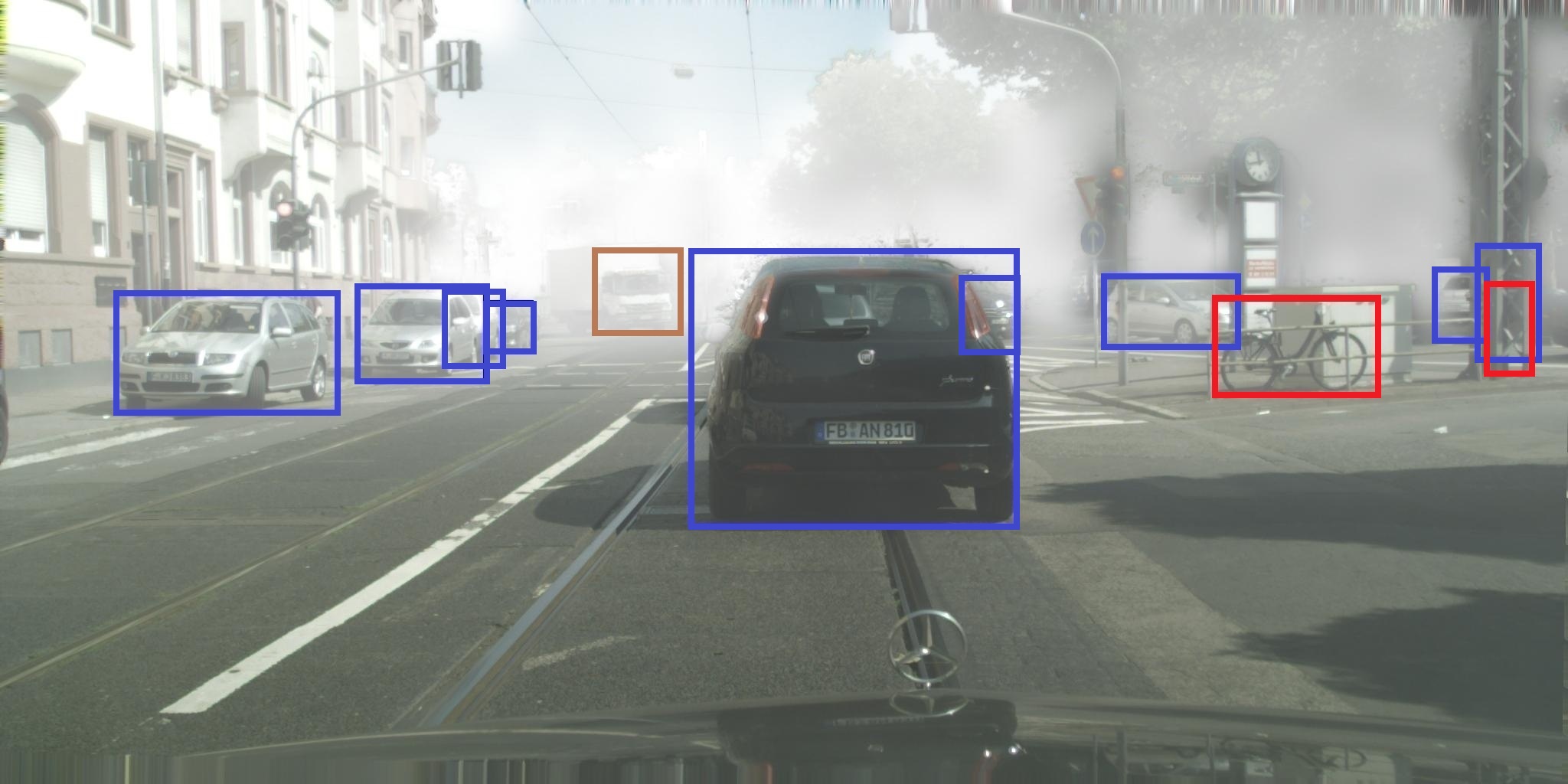}
  \end{minipage}
  }
  \hspace{0.02cm}
  \subfigure[\footnotesize Two Disentangled layers]{
  \begin{minipage}[t]{0.18\linewidth}
    \includegraphics[width=1.30in]{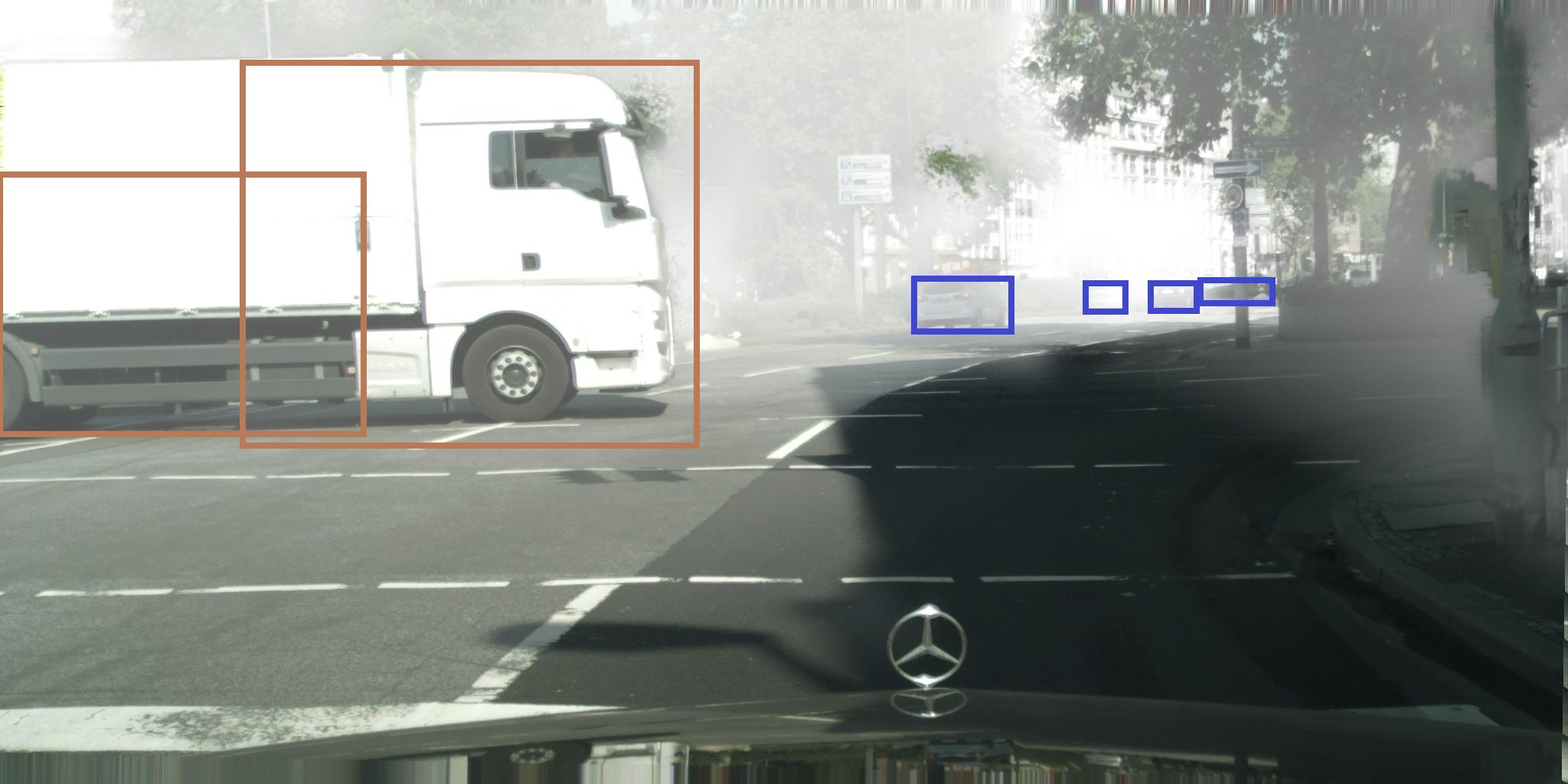}\\ \vspace{-0.1in}
    \includegraphics[width=1.30in]{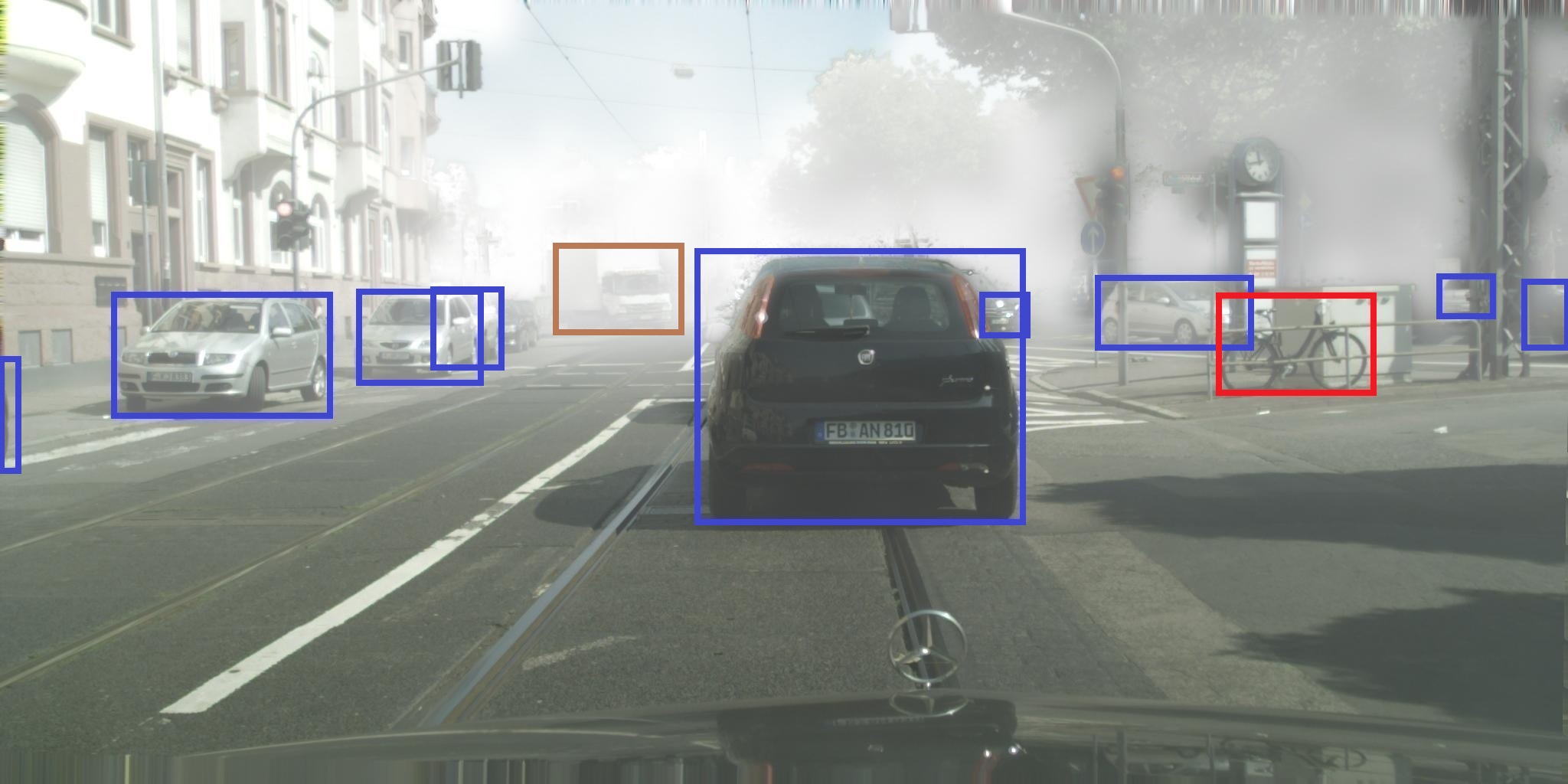}
  \end{minipage}
  }
\end{center}
\vspace{-1em}
\caption{Detection results on the ``Cityscapes $\rightarrow$ FoggyCityscapes" scene. `GT' indicates the groundtruth result. `One Disentangled layer' denotes we only use the second disentangled layer in the model. We can see that our method, i.e., using two disentangled layers, could locate and recognize objects existing in the two foggy images accurately, e.g., the \textcolor[rgb]{0.5,0,0}{truck}, \textcolor[rgb]{0,0,1}{car}, and \textcolor[rgb]{1,0,0}{bicycle}.}
\label{fig:foggycityscape}
\end{figure*}

\begin{figure*}[ht]
\vspace{-0.3em}
\begin{center}
  \subfigure[\footnotesize Raw image]{
  \begin{minipage}[t]{0.18\linewidth}
    \includegraphics[width=1.30in]{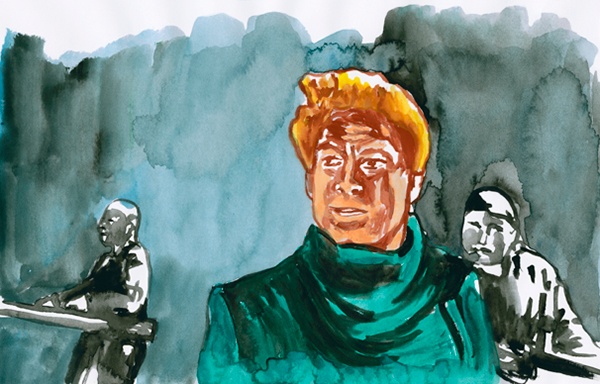}\\ \vspace{-0.1in}
    \includegraphics[width=1.30in]{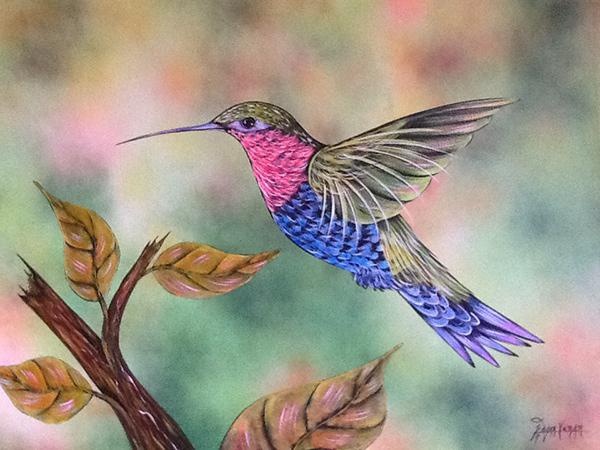}
  \end{minipage}
  }
  \hspace{0.02cm}
  \subfigure[\footnotesize GT]{
  \begin{minipage}[t]{0.18\linewidth}
    \includegraphics[width=1.30in]{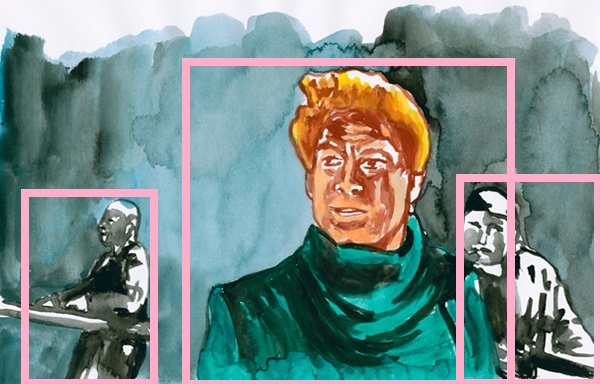}\\ \vspace{-0.1in}
    \includegraphics[width=1.30in]{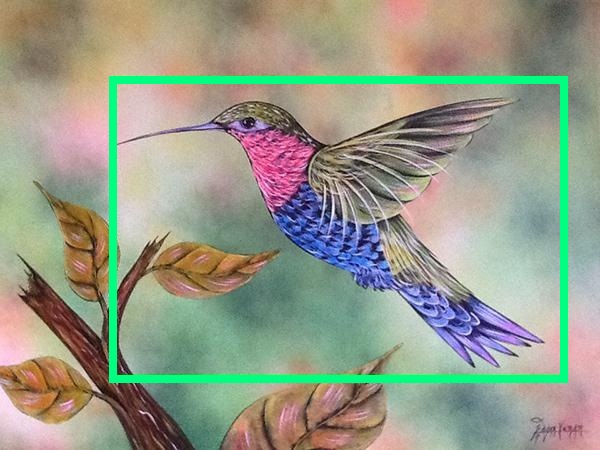}
  \end{minipage}
  }
  \hspace{0.02cm}
  \subfigure[\footnotesize SW baseline]{
  \begin{minipage}[t]{0.18\linewidth}
    \includegraphics[width=1.30in]{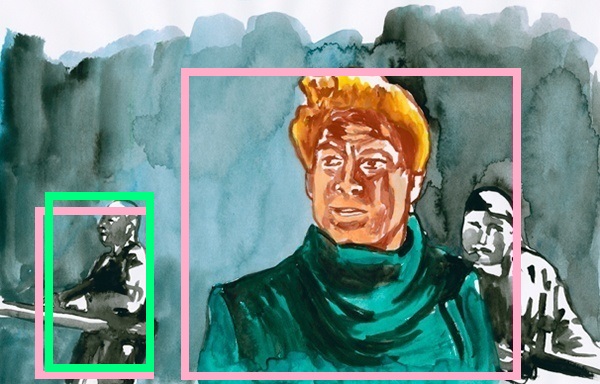}\\ \vspace{-0.1in}
    \includegraphics[width=1.30in]{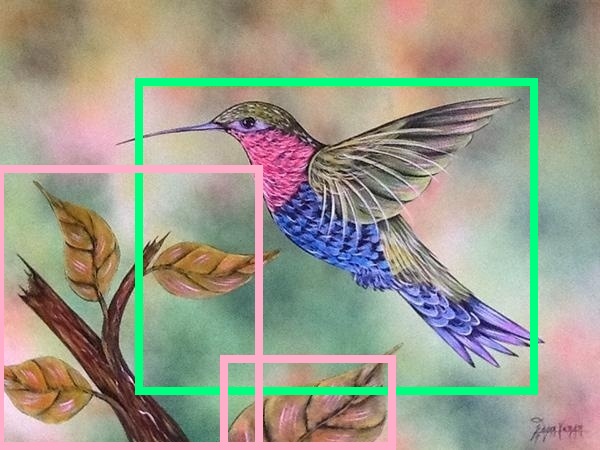}
  \end{minipage}
  }
  \hspace{0.02cm}
  \subfigure[\footnotesize One Disentangled layer]{
  \begin{minipage}[t]{0.18\linewidth}
    \includegraphics[width=1.30in]{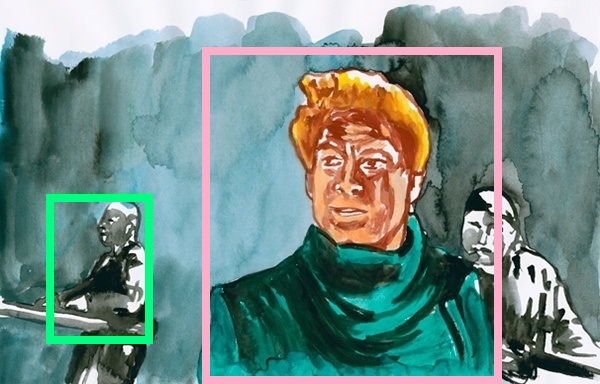}\\ \vspace{-0.1in}
    \includegraphics[width=1.30in]{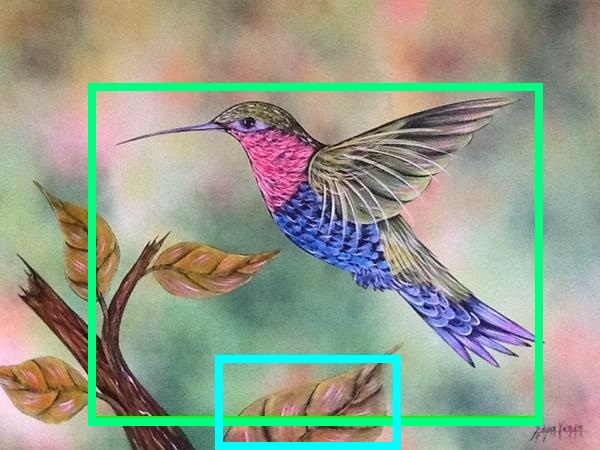}
  \end{minipage}
  }
  \hspace{0.02cm}
  \subfigure[\footnotesize Two Disentangled layers]{
  \begin{minipage}[t]{0.18\linewidth}
    \includegraphics[width=1.30in]{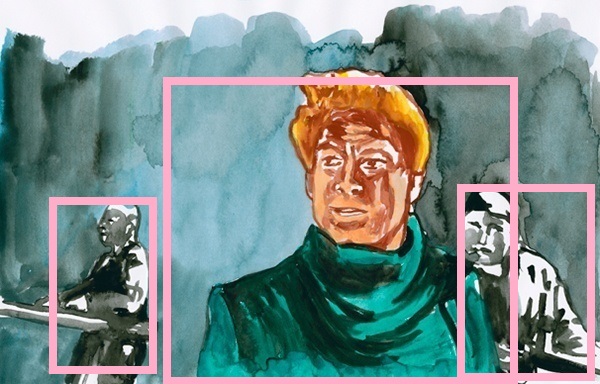}\\ \vspace{-0.1in}
    \includegraphics[width=1.30in]{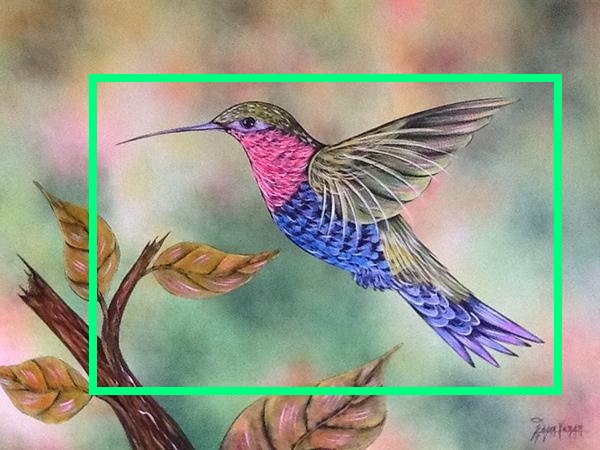}
  \end{minipage}
  }
\end{center}
\vspace{-1em}
\caption{Detection results on the ``Pascal VOC $\rightarrow$ Watercolor" scene. We can see that our method, i.e., using two disentangled layers, could locate and recognize objects existing in the two watercolor images accurately, e.g., the \textcolor[rgb]{1.0,0.61,0.61}{person}, \textcolor[rgb]{0,0.8,0}{bird}, and \textcolor[rgb]{0,0.8,0.8}{cat}.}
\label{fig:watercolor}
\end{figure*}

\begin{table}
\centering
\small
\scalebox{0.83}{
\begin{tabular}{l|cccccc|c}
\toprule[1.5pt]
Method   & bike & bird & car  & cat  & dog  & person & mAP  \\ \hline
Source Only &68.8 & 46.8 & 37.2 & 32.7 & 21.3 & 60.7   & 44.6 \\ \hline
BDC-Faster \cite{saito2019strong} &68.6 & 48.3 & 47.2 & 26.5 & 21.7 & 60.5 & 45.5 \\ \hline
DAF \cite{chen2018domain} &75.2 & 40.6 & 48.0 & 31.5 & 20.6 & 60.0 & 46.0\\ \hline \hline
SW (B) \cite{saito2019strong} & 82.3 & {\bf 55.9} & 46.5 & 32.7 & {\bf 35.5} & {\bf 66.7} & 53.3 \\
Ours & {\bf 95.8} & 54.3 & {\bf 48.3} & {\bf 42.4} & 35.1 & 65.8 & {\bf 56.9} \\
\bottomrule[1.5pt]
\end{tabular}}
\caption{Results (\%) on adaptation from Pascal to Watercolor.}\label{water}
\end{table}

\subsection{Experimental Results}

\textbf{Results on FoggyCityscapes.} Table \ref{cityscape} shows the performance of our method on the FoggyCityscapes dataset. Here, we use VGG16 and ResNet101 as the backbone of Faster-RCNN, respectively. We can see that our method outperforms all the methods in Table \ref{cityscape}. Particularly, based on the VGG16 backbone and mAP metric, our method is around 2.3\% higher than the SW baseline method \cite{saito2019strong}. Compared with RLDA \cite{khodabandeh2019robust} using InceptionV2 \cite{szegedy2016rethinking} as the strong backbone, our method still outperforms it. These all show our method is effective. Moreover, employing the backbone of ResNet101 could improve the performance of our method significantly. This shows our method is more effective with a better backbone. Fig. \ref{fig:foggycityscape} shows two detection examples. Compared with the raw images, for object detection, the foggy scene is much more challenging. Meanwhile, compared with the SW method, our method could locate and recognize objects existing in the two images accurately. Particularly, regardless of distance, our method could locate and discriminate the truck accurately. These further demonstrate the effectiveness of our method.

\textbf{Results on Watercolor and Clipart.} Table \ref{water} and \ref{clipart} separately show the performance of our method on Watercolor and Clipart dataset. Here, we all use ResNet101 as the backbone of Faster-RCNN. For Watercolor scene, our method is 3.6\% higher than the SW method. Particularly, for the class of bike, our method outperforms SW by around 13\%. This shows our method is effective for the task of DAOD. Fig. \ref{fig:watercolor} shows two examples of Watercolor. We can see that our method could locate and recognize the classes of person and bird accurately. This further shows that our disentangled method indeed alleviates the problem of domain-shift and improves the detection performance.

\begin{figure*}[ht]
\begin{center}
  \subfigure[\footnotesize GT]{
  \begin{minipage}[t]{0.12\linewidth}
    \includegraphics[width=0.90in,height=0.7in]{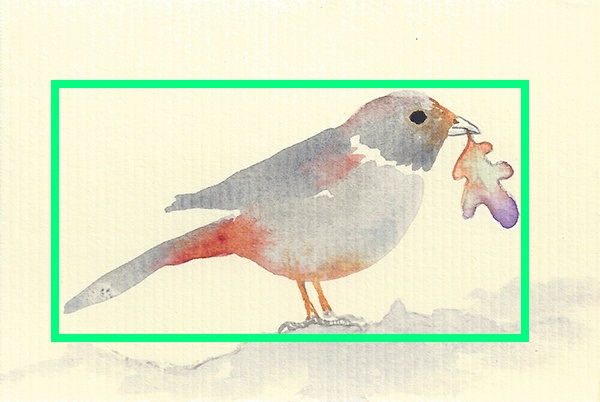}\\ \vspace{-0.1in}
    \includegraphics[width=0.90in,height=0.9in]{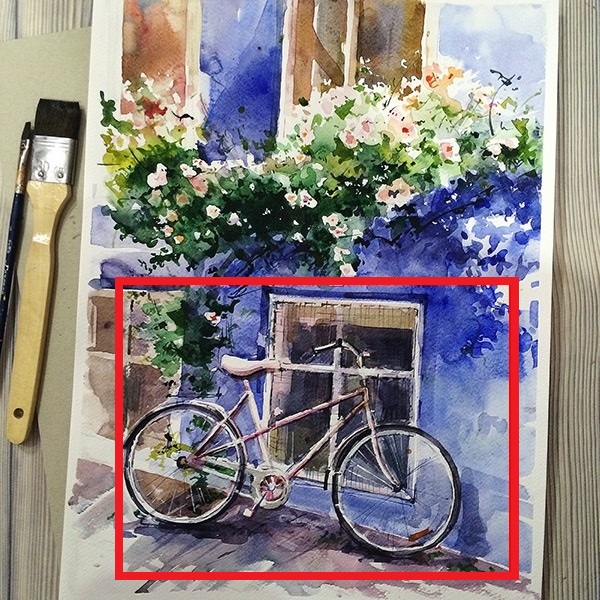}
  \end{minipage}
  }
  \hspace{0.02cm}
  \subfigure[\footnotesize O-Base]{
  \begin{minipage}[t]{0.12\linewidth}
    \includegraphics[width=0.90in,height=0.7in]{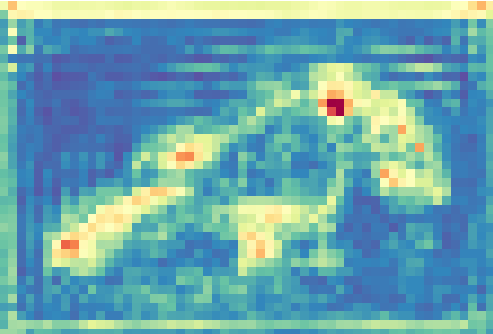}\\ \vspace{-0.1in}
    \includegraphics[width=0.90in,height=0.9in]{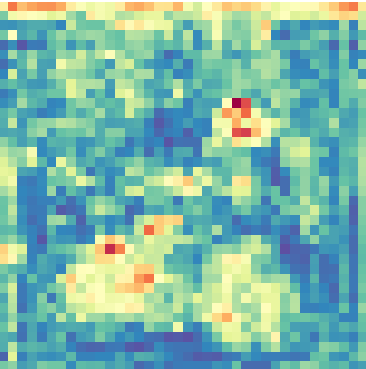}
  \end{minipage}
  }
  \hspace{-0.01em}
  \subfigure[\footnotesize P-Base]{
  \begin{minipage}[t]{0.12\linewidth}
    \includegraphics[width=0.90in,height=0.7in]{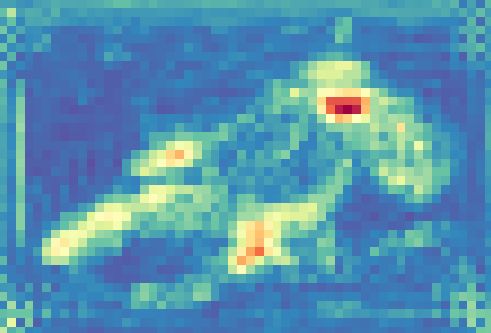}\\ \vspace{-0.1in}
    \includegraphics[width=0.90in,height=0.9in]{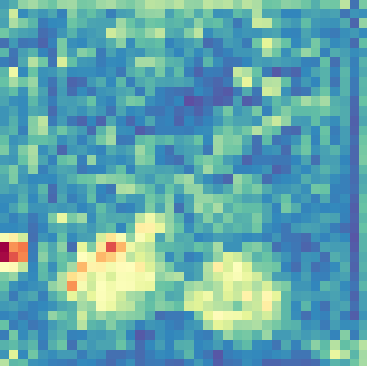}
  \end{minipage}
  }
  \hspace{0.04cm}
  \subfigure[\footnotesize O-DIR]{
  \begin{minipage}[t]{0.12\linewidth}
    \includegraphics[width=0.90in,height=0.7in]{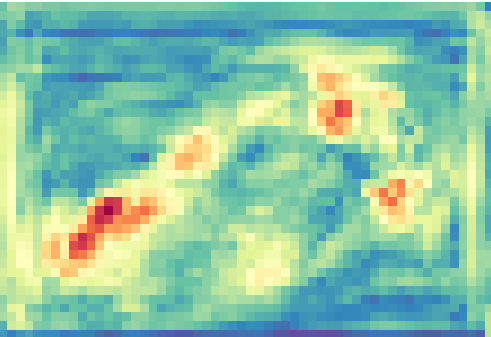}\\ \vspace{-0.1in}
    \includegraphics[width=0.90in,height=0.9in]{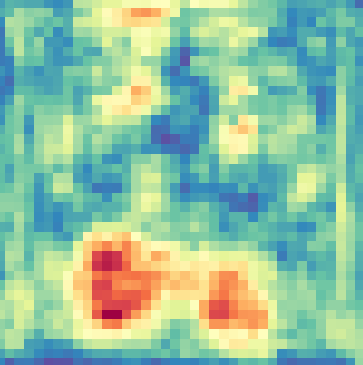}
  \end{minipage}
  }
  \hspace{-0.01em}
  \subfigure[\footnotesize P-DIR]{
  \begin{minipage}[t]{0.12\linewidth}
    \includegraphics[width=0.90in,height=0.7in]{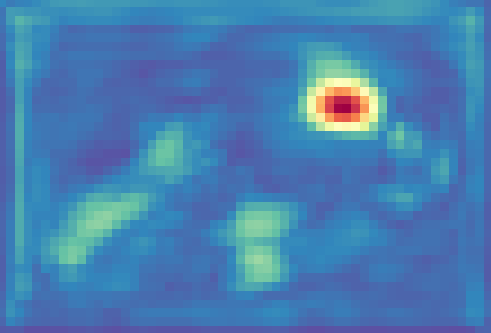}\\ \vspace{-0.1in}
    \includegraphics[width=0.90in,height=0.9in]{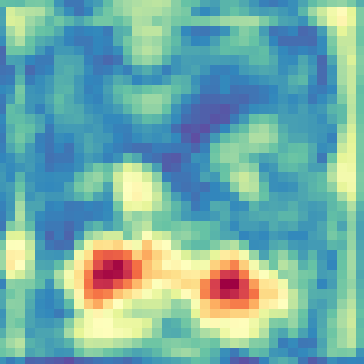}
  \end{minipage}
  }
  \hspace{0.04cm}
  \subfigure[\footnotesize O-DSR]{
  \begin{minipage}[t]{0.12\linewidth}
    \includegraphics[width=0.90in,height=0.7in]{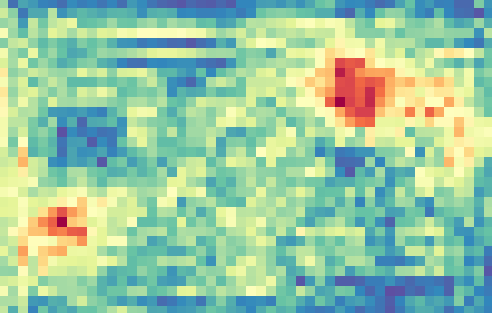}\\ \vspace{-0.1in}
    \includegraphics[width=0.90in,height=0.9in]{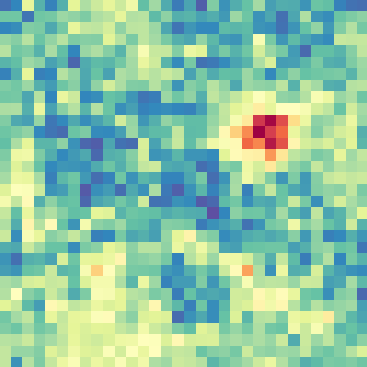}
  \end{minipage}
  }
  \hspace{-0.01em}
  \subfigure[\footnotesize P-DSR]{
  \begin{minipage}[t]{0.12\linewidth}
    \includegraphics[width=0.90in,height=0.7in]{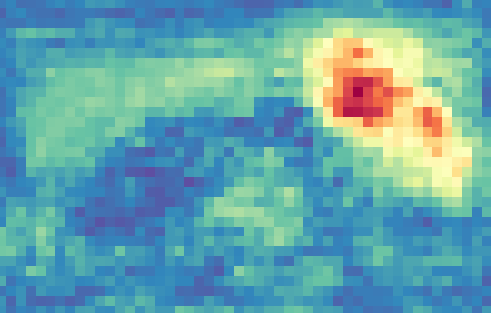}\\ \vspace{-0.1in}
    \includegraphics[width=0.90in,height=0.9in]{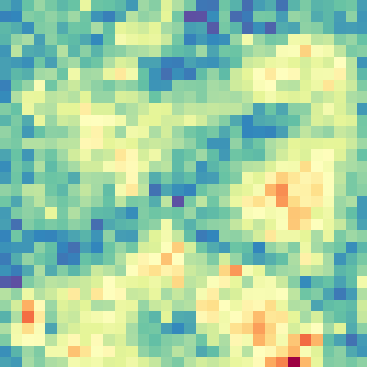}
  \end{minipage}
  }
\end{center}
\vspace{-1em}
\caption{Visualization of feature maps of the second disentangled layer. Here, `O-DIR' ($F_{di}^{2}$) and `O-DSR' ($F_{ds}^{2}$) indicate we only use the second disentangled layer to extract DIR and DSR based on `O-Base' ($F_{b}^{2}$) and do not use the first disentangled layer. `P-DIR' ($F_{di}^{2}$) and `P-DSR' ($F_{ds}^{2}$) indicate we use the progressive method to extract DIR and DSR based on `P-Base' ($F_{b}^{2}$). For each feature map, the channels corresponding to the maximum value are selected for visualization. For `O-DIR' and `P-DIR', the bright regions indicate the presentence of object-relevant content. For `O-DSR' and `P-DSR', the bright regions indicate the presentence of domain-specific information.}
\label{fig:visualization}
\end{figure*}

\begin{table*}
\small
\begin{center}
\scalebox{0.98}{
\begin{tabular}{@{\hskip3pt}l@{\hskip2pt}|@{\hskip2pt}c@{\hskip3pt}c@{\hskip3pt}c@{\hskip3pt}c@{\hskip3pt}c@{\hskip3pt}c@{\hskip3pt}c@{\hskip3pt}c@{\hskip3pt}c@{\hskip3pt}c@{\hskip3pt}c@{\hskip3pt}c@{\hskip3pt}c@{\hskip3pt}c@{\hskip3pt}c@{\hskip3pt}c@{\hskip3pt}c@{\hskip3pt}c@{\hskip3pt}c@{\hskip3pt}c@{\hskip2pt}|@{\hskip2pt}c@{\hskip2pt}}
\toprule[1.5pt]
Method          & aero & bike & bird & boat & bottle & bus & car & cat & chair & cow & table & dog & horse & mbike & person & plant & sheep & sofa & train & tv  &  mAP  \\ \hline
Source Only        & 35.6&	52.5&	24.3&	23.0&	20.0&	43.9&	32.8&	10.7&	30.6&	11.7&	13.8&	6.0&	36.8&	45.9&	48.7&	41.9&	{\bf 16.5}&	7.3&	22.9&	32.0&	27.8 \\ \hline
BDC-Faster \cite{saito2019strong} & 20.2 & 46.4 & 20.4 & 19.3 & 18.7 & 41.3 & 26.5 & 6.4 & 33.2 & 11.7 & 26.0 & 1.7 & 36.6 & 41.5 & 37.7 & 44.5 & 10.6 & 20.4 & 33.3 & 15.5 & 25.6 \\ \hline
DAF \cite{chen2018domain} & 15.0 & 34.6 & 12.4 & 11.9 & 19.8 & 21.1 & 23.2 & 3.1 & 22.1 & 26.3 & 10.0 & 10.0 & 19.6 & 39.4 & 34.6 & 29.3 & 1.0 & 17.1 & 19.7 & 24.8 & 19.8 \\ \hline \hline
SW (B) \cite{saito2019strong} & 26.2 & 48.5 & 32.6 & {\bf 33.7} & 38.5 & 54.3 & 37.1 & {\bf 18.6} & 34.8 & {\bf 58.3} & 17.0 & 12.5 & 33.8 & 65.5 & 61.6 & {\bf 52.0} & 9.3 & 24.9 & 54.1 & {\bf 49.1} & 38.1 \\
Ours & {\bf 41.5} & {\bf 52.7} & {\bf 34.5} & 28.1 & {\bf 43.7} & {\bf 58.5} & {\bf 41.8} & 15.3 & {\bf 40.1} & 54.4 & {\bf 26.7} & {\bf 28.5} & {\bf 37.7} & {\bf 75.4} & {\bf 63.7} & 48.7 & {\bf 16.5} & {\bf 30.8} & {\bf 54.5} & 48.7 & {\bf 42.1} \\ \bottomrule[1.5pt]
\end{tabular}}
\caption{Results (\%) on adaptation from Pascal VOC to Clipart. Here, we use ResNet101 as the backbone of Faster-RCNN.}\label{clipart}
\end{center}
\end{table*}

As for Clipart scene which involves more classes than the other two datasets, our method outperforms SW by 4.0\%, in terms of the mAP metric. Meanwhile, in Table \ref{clipart}, we can see that our method outperforms the baseline method in multiple categories significantly. For example, for the aeroplane and dog class, our method is around 15\% and 16\% higher than the SW method. These all demonstrate the good performance of our method.

\subsection{Ablation Analysis}

In this section, we will make some ablation analysis on our method. Table \ref{ablation} shows the ablation results. Here, `C $\rightarrow$ F' and `V $\rightarrow$ W' separately indicate the adaptation from Cityscapes to FoggyCityscapes and the adaptation from Pascal VOC to Watercolor. And for the `C $\rightarrow$ F' case, we use VGG16 as the backbone. For the `V $\rightarrow$ W' case, we use ResNet101 as the backbone. `OW' indicates we integrate all loss functions existing in our method and use one training stage. `1st', `2nd', and `3rd' indicate we use the first training stage of Algorithm \ref{alg_DAOD}, the first two training stages of Algorithm \ref{alg_DAOD}, and the three training stages to optimize our model, respectively. For our progressive method (Two layers), we can see that the three-stage training mechanism is effective. For example, for the `C $\rightarrow$ F' case, the performance is improved from 33.6\% to 36.6\%. Meanwhile, we can see that from the first training stage to the third stage, the performance is improved continuously. This shows that for the disentangled learning, the stage of feature separation and feature reconstruction is necessary. Using these two stages does enhance the disentanglement and improve the detection performance. Besides, we can also see that the relation-consistency loss (RC) improves the performance of our method significantly. For example, for the `V $\rightarrow$ W' scene, the performance is improved from 55.2\% to 56.9\%. This demonstrates the relation-consistency loss helps strengthen the ability of disentanglement.

To further verify the effectiveness of the progressive method, we make a comparison with the method of only using the second disentangled layer (One layer). We can see from Table \ref{ablation} that our progressive method improves the detection performance significantly, e.g., for the `C $\rightarrow$ F' case, the performance is improved from 34.1\% to 36.6\%. This shows that using the progressive mechanism is indeed helpful for obtaining a better disentangled representation. Besides, in Fig. \ref{fig:foggycityscape} and \ref{fig:watercolor}, we can see that compared with One layer method, employing two disentangled layers does improve the accuracy of location and recognition. Particularly, taking the first image in Fig. \ref{fig:watercolor} as an example, our method accurately locates and classifies the three persons existing in the watercolor image. This further demonstrates the good performance of our method.

\begin{table}
\centering
\scalebox{0.87}{
\begin{tabular}{l|ccccc|c|c}
\hline
Method  & OW & 1st & 2nd & 3rd & RC & C $\rightarrow$ F & V $\rightarrow$ W \\ \hline
Two layers &$\checkmark$ & & & &$\checkmark$ & 34.1\% & 52.9\% \\
Two layers & &$\checkmark$ & & & & 33.6\% & 53.5\% \\
Two layers & & &$\checkmark$ & &$\checkmark$ &35.3\% & 55.3\% \\
Two layers & & & &$\checkmark$ & & 35.5\% & 55.2\% \\
Two layers & & & &$\checkmark$ &$\checkmark$ & \textbf{36.6\%} & \textbf{56.9\%} \\ \hline
\hline
One layer & & & &$\checkmark$ &$\checkmark$ & 34.1\% & 54.6\% \\
Two layers & & & &$\checkmark$ &$\checkmark$ & \textbf{36.6\%} & \textbf{56.9\%} \\ \hline
\end{tabular}}
\caption{Ablation analysis of the proposed progressive disentanglement. Here, we use mAP as the metric.}\label{ablation}
\end{table}

\subsection{Visualization Analysis}

\begin{figure*}[p]
\begin{center}
  \subfigure[\footnotesize GT]{
  \begin{minipage}[t]{0.18\linewidth}
    \includegraphics[width=1.25in,height=1.0in]{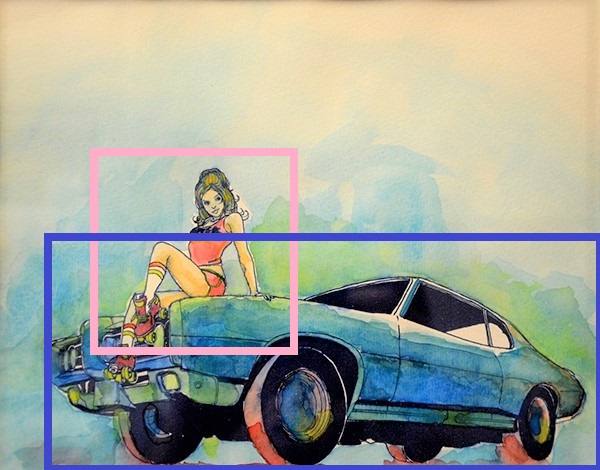} \vspace{-0.1in} \\
    \includegraphics[width=1.25in,height=1.0in]{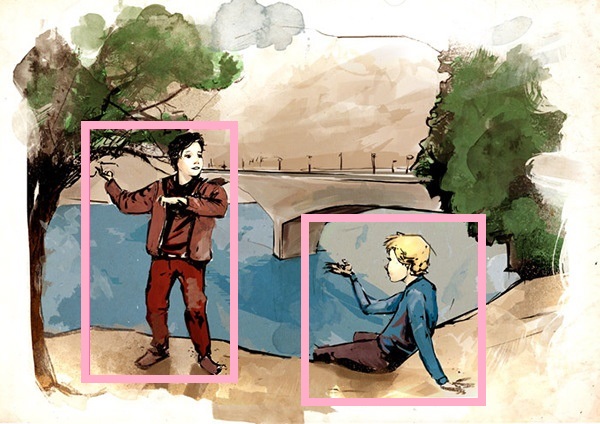} \vspace{-0.1in} \\
    \includegraphics[width=1.25in,height=1.5in]{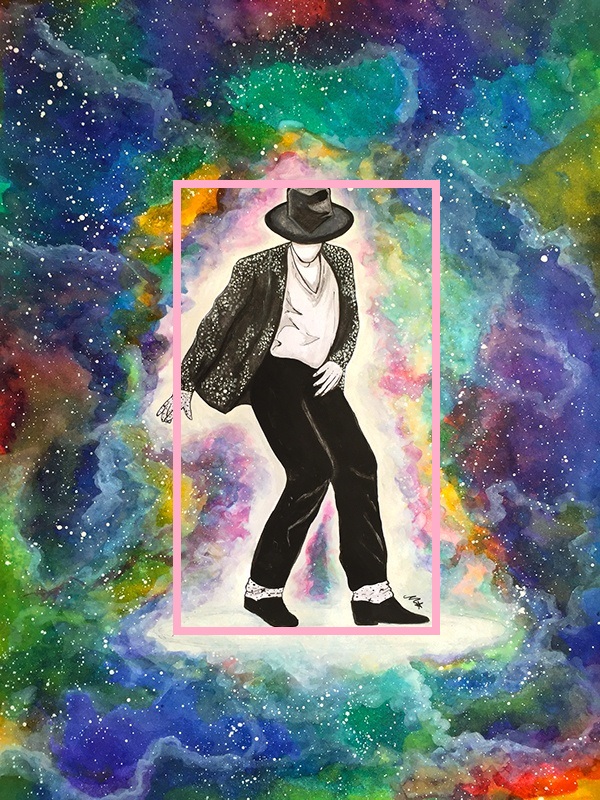} \vspace{-0.1in} \\
    \includegraphics[width=1.25in,height=1.0in]{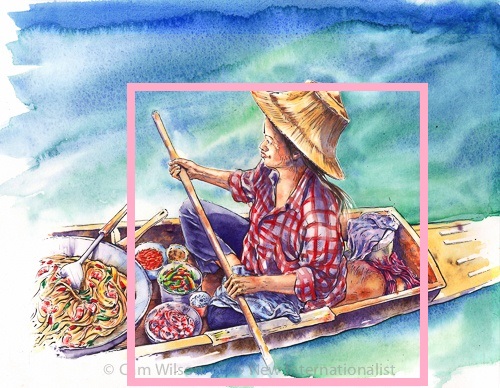} \vspace{-0.1in} \\
    \includegraphics[width=1.25in,height=1.0in]{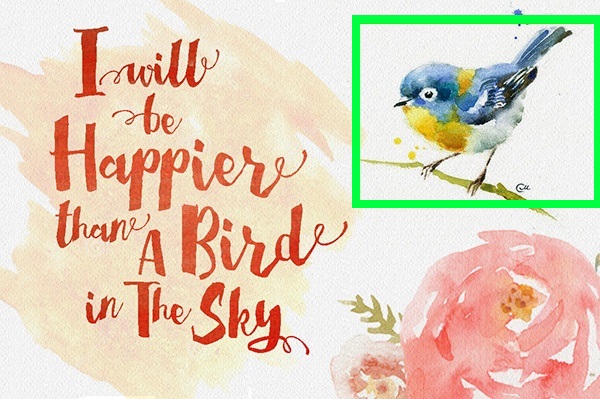} \vspace{-0.1in} \\
    \includegraphics[width=1.25in,height=1.0in]{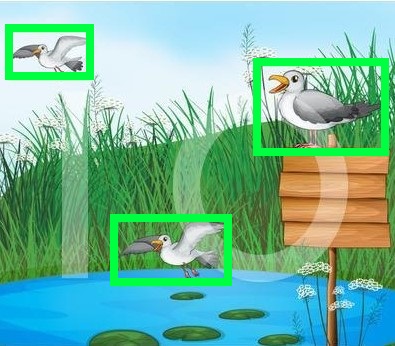} \vspace{-0.1in} \\
    \includegraphics[width=1.25in,height=1.3in]{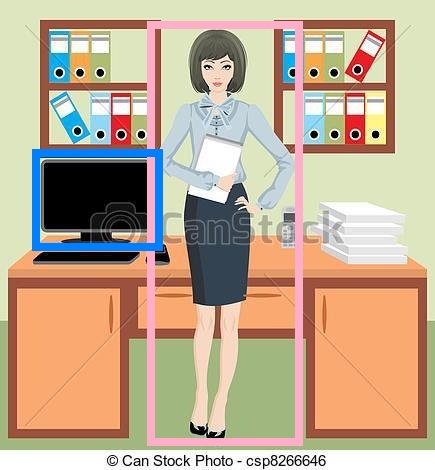}
  \end{minipage}
  }
  \subfigure[\footnotesize Detection Results]{
  \begin{minipage}[t]{0.18\linewidth}
    \includegraphics[width=1.25in,height=1.0in]{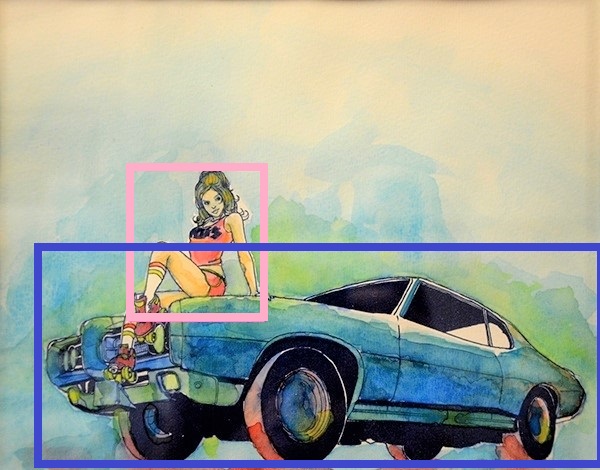} \vspace{-0.1in} \\
    \includegraphics[width=1.25in,height=1.0in]{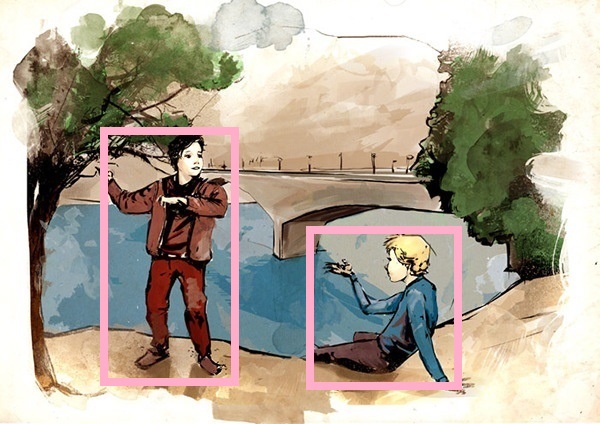} \vspace{-0.1in} \\
    \includegraphics[width=1.25in,height=1.5in]{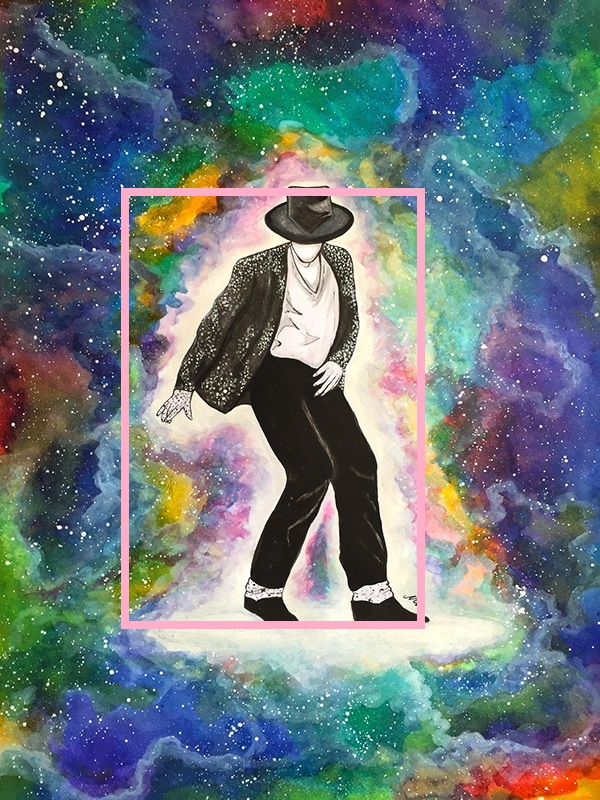} \vspace{-0.1in} \\
    \includegraphics[width=1.25in,height=1.0in]{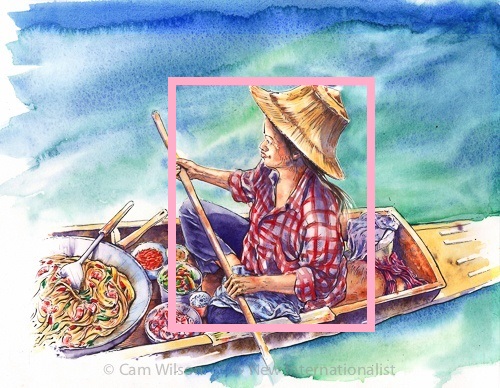} \vspace{-0.1in} \\
    \includegraphics[width=1.25in,height=1.0in]{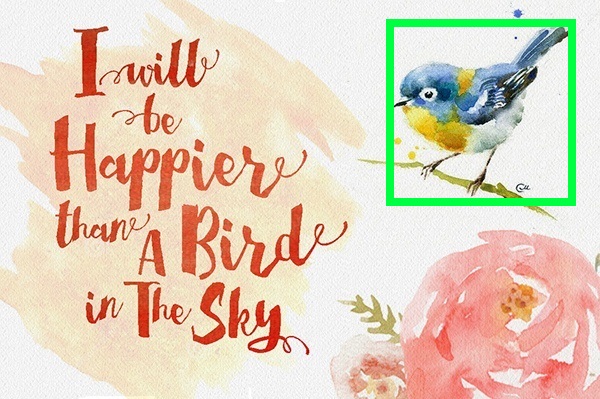} \vspace{-0.1in} \\
    \includegraphics[width=1.25in,height=1.0in]{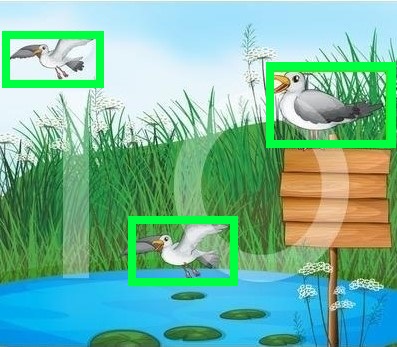} \vspace{-0.1in} \\
    \includegraphics[width=1.25in,height=1.3in]{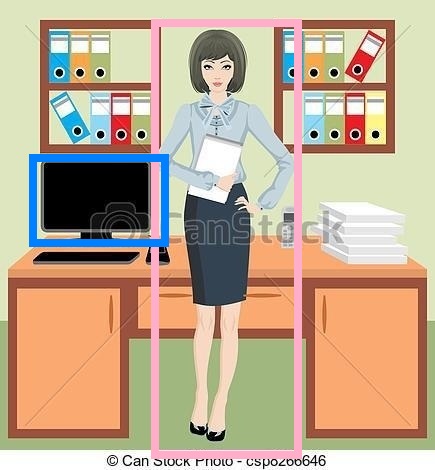}
  \end{minipage}
  }
  \subfigure[\footnotesize P-Base]{
  \begin{minipage}[t]{0.18\linewidth}
    \includegraphics[width=1.25in,height=1.0in]{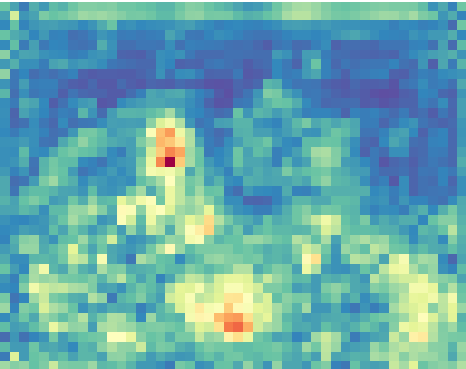} \vspace{-0.1in} \\
    \includegraphics[width=1.25in,height=1.0in]{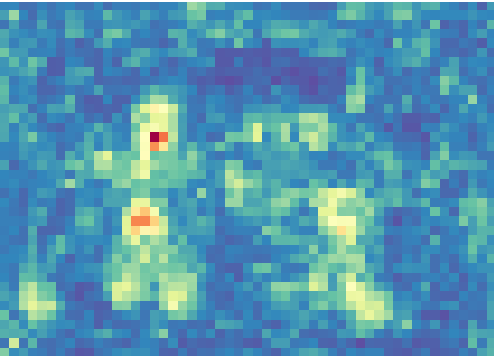} \vspace{-0.1in} \\
    \includegraphics[width=1.25in,height=1.5in]{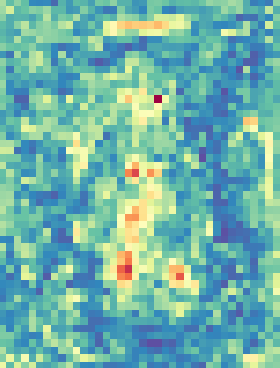} \vspace{-0.1in} \\
    \includegraphics[width=1.25in,height=1.0in]{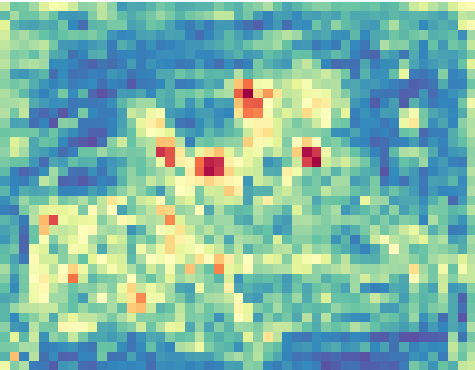} \vspace{-0.1in} \\
    \includegraphics[width=1.25in,height=1.0in]{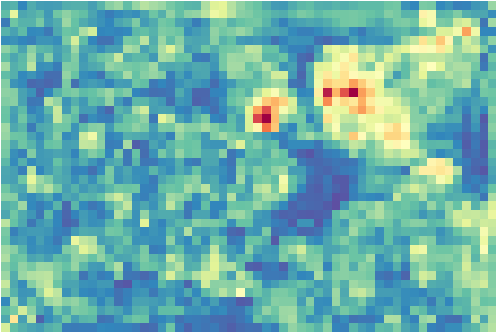} \vspace{-0.1in} \\
    \includegraphics[width=1.25in,height=1.0in]{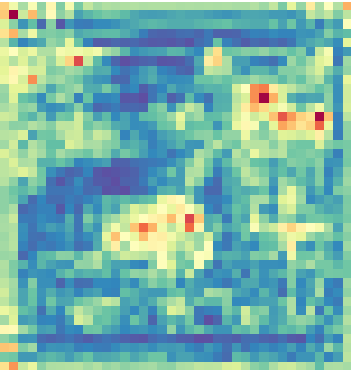} \vspace{-0.1in} \\
    \includegraphics[width=1.25in,height=1.3in]{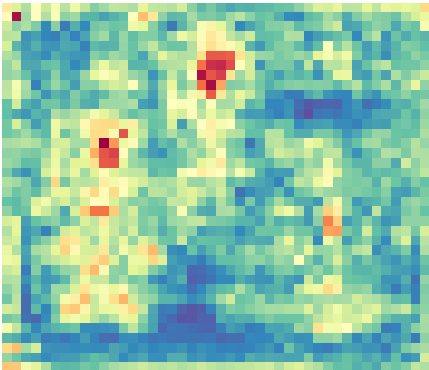}
  \end{minipage}
  }
  \subfigure[\footnotesize P-DIR]{
  \begin{minipage}[t]{0.18\linewidth}
    \includegraphics[width=1.25in,height=1.0in]{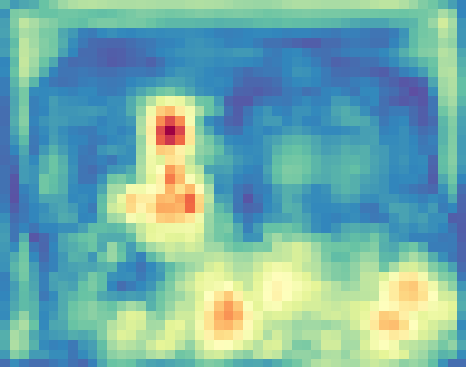} \vspace{-0.1in} \\
    \includegraphics[width=1.25in,height=1.0in]{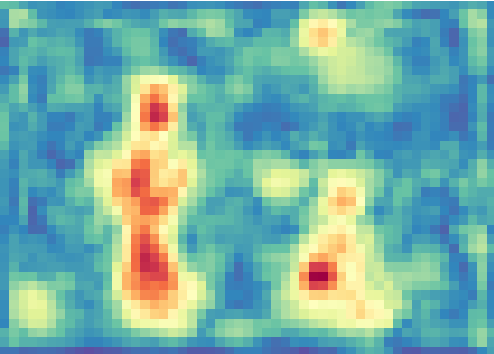} \vspace{-0.1in} \\
    \includegraphics[width=1.25in,height=1.5in]{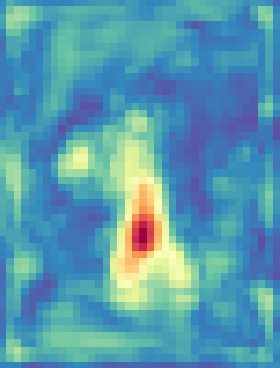} \vspace{-0.1in} \\
    \includegraphics[width=1.25in,height=1.0in]{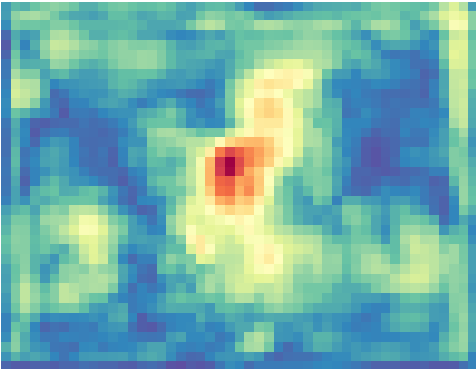} \vspace{-0.1in} \\
    \includegraphics[width=1.25in,height=1.0in]{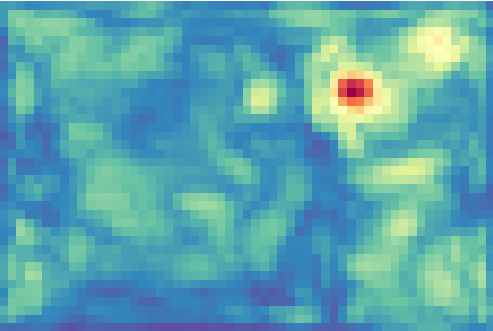} \vspace{-0.1in} \\
    \includegraphics[width=1.25in,height=1.0in]{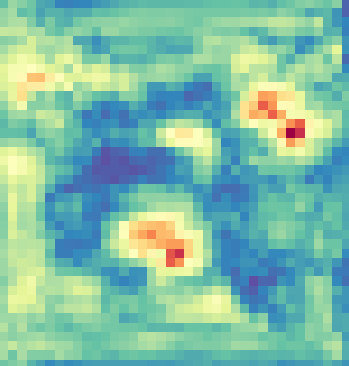} \vspace{-0.1in} \\
    \includegraphics[width=1.25in,height=1.3in]{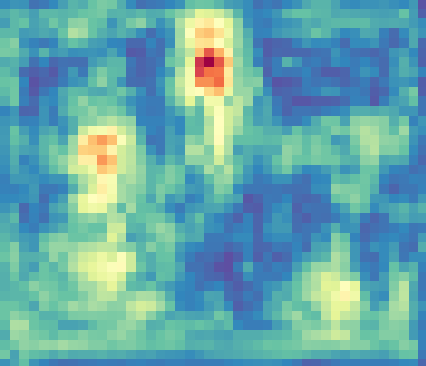}
  \end{minipage}
  }
  \subfigure[\footnotesize P-DSR]{
  \begin{minipage}[t]{0.18\linewidth}
    \includegraphics[width=1.25in,height=1.0in]{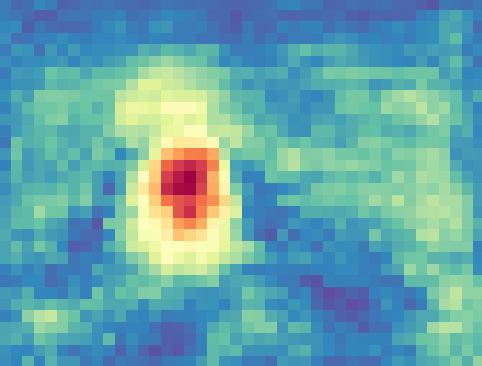} \vspace{-0.1in} \\
    \includegraphics[width=1.25in,height=1.0in]{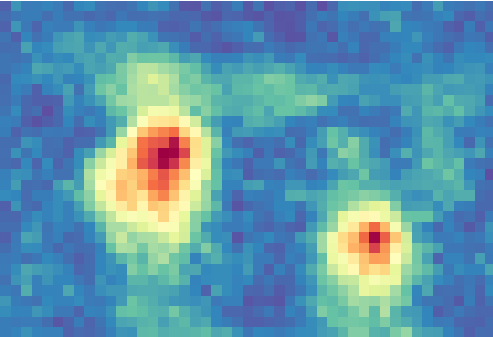} \vspace{-0.1in} \\
    \includegraphics[width=1.25in,height=1.5in]{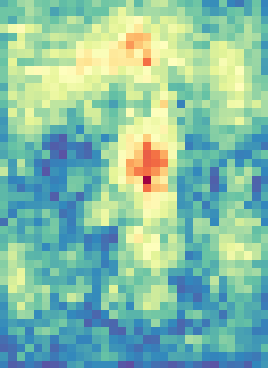} \vspace{-0.1in} \\
    \includegraphics[width=1.25in,height=1.0in]{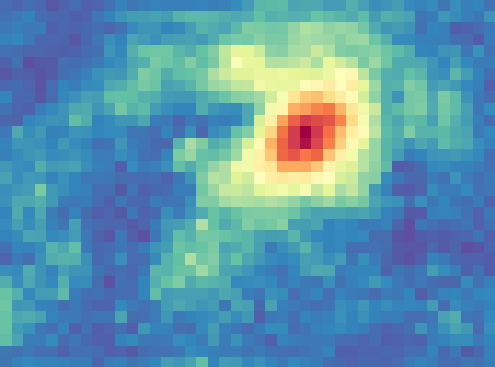} \vspace{-0.1in} \\
    \includegraphics[width=1.25in,height=1.0in]{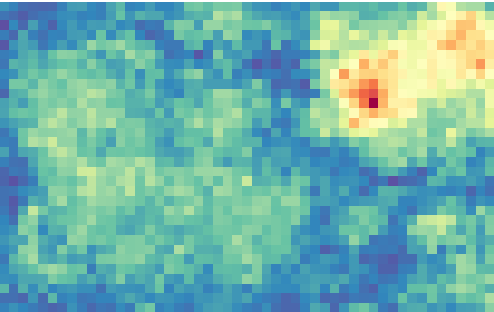} \vspace{-0.1in} \\
    \includegraphics[width=1.25in,height=1.0in]{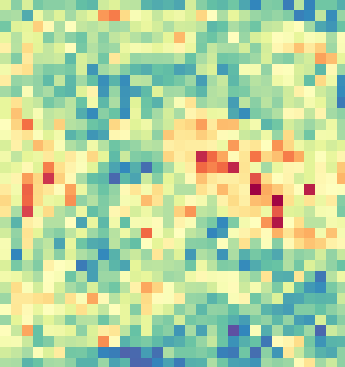} \vspace{-0.1in} \\
    \includegraphics[width=1.25in,height=1.3in]{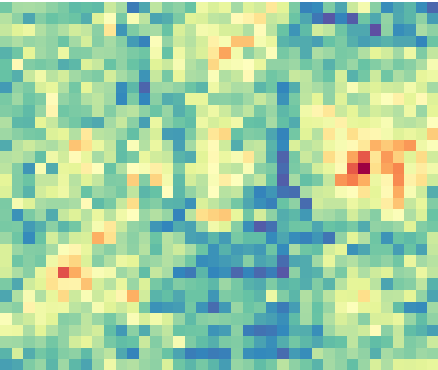}
  \end{minipage}
  }
\end{center}
\vspace{-1.0em}
\caption{Visualization of feature maps of the second disentangled layer. Here, we use the progressive disentangled method to extract DIR and DSR. `Base' indicates the feature map used for disentanglement. The examples of the first five rows are from the `Pascal VOC $\rightarrow$ Watercolor' scene. The examples of the last two rows are from the `Pascal VOC $\rightarrow$ Clipart' scene.}
\label{visual:water}
\end{figure*}

In Fig. \ref{fig:visualization}, taking two watercolor images as examples, a visualization analysis is made to show the learned disentangled representations. We can see both the method of only using the second disentangled layer and the progressive method could learn good disentangled representations. Particularly, compared with the `O-Base' and `P-Base' used for disentanglement, the learned DIR and DSR separately contain much stronger object-relevant information and domain-specific information. These results demonstrate that our method can successfully learn disentangled representations. Besides, compared with `O-Base', `P-Base' contains much less domain-specific information, e.g., the background information in the first image and the color wall in the second image. This shows the first disentangled layer indeed enhances the domain-invariant information. Meanwhile, compared with `O-DIR', our progressive method can extract a better DIR. Particularly, for these two images, `P-DIR' is much smoother and contains much less domain-specific information. For example, the leaf and background information in the first image, and the flowers in the second image are much less in `P-DIR', which is helpful for the location and recognition of objects. These all show our progressive method really owns the disentanglement ability and learns better instance-invariant features that lead to a better detection performance. More visualization examples can be seen in Fig. \ref{visual:water}.

\section{Conclusion}

In this paper, we focus on obtaining the instance-invariant features for solving domain adaptive object detection. A progressive disentangled framework is first proposed to decompose domain-invariant and domain-specific features. Then, the instance-invariant features are extracted based on the domain-invariant features, which could alleviate the problem of domain-shift. Finally, we propose a three-stage training mechanism to enhance the disentanglement. In the experiment, our method achieves a new state-of-the-art performance on three domain-shift scenes.

{\small
\bibliographystyle{ieee_fullname}
\bibliography{egbib}

\begin{thebibliography}{10}\itemsep=-1pt

\bibitem{belghazi2018mine}
Mohamed~Ishmael Belghazi, Aristide Baratin, Sai Rajeswar, Sherjil Ozair, Yoshua
  Bengio, Aaron Courville, and R~Devon Hjelm.
\newblock Mine: mutual information neural estimation.
\newblock {\em ICML}, 2018.

\bibitem{bengio2013representation}
Yoshua Bengio, Aaron Courville, and Pascal Vincent.
\newblock Representation learning: A review and new perspectives.
\newblock {\em IEEE transactions on pattern analysis and machine intelligence},
  35(8):1798--1828, 2013.

\bibitem{bousmalis2017unsupervised}
Konstantinos Bousmalis, Nathan Silberman, David Dohan, Dumitru Erhan, and Dilip
  Krishnan.
\newblock Unsupervised pixel-level domain adaptation with generative
  adversarial networks.
\newblock In {\em Proceedings of the IEEE conference on computer vision and
  pattern recognition}, pages 3722--3731, 2017.

\bibitem{cai2019exploring}
Qi Cai, Yingwei Pan, Chong-Wah Ngo, Xinmei Tian, Lingyu Duan, and Ting Yao.
\newblock Exploring object relation in mean teacher for cross-domain detection.
\newblock In {\em Proceedings of the IEEE Conference on Computer Vision and
  Pattern Recognition}, pages 11457--11466, 2019.

\bibitem{chen2018domain}
Yuhua Chen, Wen Li, Christos Sakaridis, Dengxin Dai, and Luc Van~Gool.
\newblock Domain adaptive faster r-cnn for object detection in the wild.
\newblock In {\em Proceedings of the IEEE conference on computer vision and
  pattern recognition}, pages 3339--3348, 2018.

\bibitem{cicek2019unsupervised}
Safa Cicek and Stefano Soatto.
\newblock Unsupervised domain adaptation via regularized conditional alignment.
\newblock {\em arXiv preprint arXiv:1905.10885}, 2019.

\bibitem{cordts2016cityscapes}
Marius Cordts, Mohamed Omran, Sebastian Ramos, Timo Rehfeld, Markus Enzweiler,
  Rodrigo Benenson, Uwe Franke, Stefan Roth, and Bernt Schiele.
\newblock The cityscapes dataset for semantic urban scene understanding.
\newblock In {\em Proceedings of the IEEE conference on computer vision and
  pattern recognition}, pages 3213--3223, 2016.

\bibitem{do2019theory}
Kien Do and Truyen Tran.
\newblock Theory and evaluation metrics for learning disentangled
  representations.
\newblock {\em arXiv preprint arXiv:1908.09961}, 2019.

\bibitem{everingham2010pascal}
Mark Everingham, Luc Van~Gool, Christopher~KI Williams, John Winn, and Andrew
  Zisserman.
\newblock The pascal visual object classes (voc) challenge.
\newblock {\em International journal of computer vision}, 88(2):303--338, 2010.

\bibitem{ganin2014unsupervised}
Yaroslav Ganin and Victor Lempitsky.
\newblock Unsupervised domain adaptation by backpropagation.
\newblock {\em arXiv preprint arXiv:1409.7495}, 2014.

\bibitem{girshick2015fast}
Ross Girshick.
\newblock Fast r-cnn.
\newblock In {\em Proceedings of the IEEE international conference on computer
  vision}, pages 1440--1448, 2015.

\bibitem{gong2019dlow}
Rui Gong, Wen Li, Yuhua Chen, and Luc~Van Gool.
\newblock Dlow: Domain flow for adaptation and generalization.
\newblock In {\em Proceedings of the IEEE Conference on Computer Vision and
  Pattern Recognition}, pages 2477--2486, 2019.

\bibitem{he2017mask}
Kaiming He, Georgia Gkioxari, Piotr Doll{\'a}r, and Ross Girshick.
\newblock Mask r-cnn.
\newblock In {\em Proceedings of the IEEE international conference on computer
  vision}, pages 2961--2969, 2017.

\bibitem{he2019multi}
Zhenwei He and Lei Zhang.
\newblock Multi-adversarial faster-rcnn for unrestricted object detection.
\newblock {\em arXiv preprint arXiv:1907.10343}, 2019.

\bibitem{hu2018relation}
Han Hu, Jiayuan Gu, Zheng Zhang, Jifeng Dai, and Yichen Wei.
\newblock Relation networks for object detection.
\newblock In {\em Proceedings of the IEEE Conference on Computer Vision and
  Pattern Recognition}, pages 3588--3597, 2018.

\bibitem{huang2018multimodal}
Xun Huang, Ming-Yu Liu, Serge Belongie, and Jan Kautz.
\newblock Multimodal unsupervised image-to-image translation.
\newblock In {\em Proceedings of the European Conference on Computer Vision
  (ECCV)}, pages 172--189, 2018.

\bibitem{inoue2018cross}
Naoto Inoue, Ryosuke Furuta, Toshihiko Yamasaki, and Kiyoharu Aizawa.
\newblock Cross-domain weakly-supervised object detection through progressive
  domain adaptation.
\newblock In {\em Proceedings of the IEEE conference on computer vision and
  pattern recognition}, pages 5001--5009, 2018.

\bibitem{Jiang2018Graph}
Bo Jiang, Ziyan Zhang, Doudou Lin, and Jin Tang.
\newblock Graph learning-convolutional networks.
\newblock {\em ICML}, 2019.

\bibitem{khodabandeh2019robust}
Mehran Khodabandeh, Arash Vahdat, Mani Ranjbar, and William~G Macready.
\newblock A robust learning approach to domain adaptive object detection.
\newblock {\em ICCV}, 2019.

\bibitem{Kim2019SelfTrainingAA}
Seunghyeon Kim, Jaehoon Choi, Taekyung Kim, and Changick Kim.
\newblock Self-training and adversarial background regularization for
  unsupervised domain adaptive one-stage object detection.
\newblock {\em ArXiv}, abs/1909.00597, 2019.

\bibitem{kim2019diversify}
Taekyung Kim, Minki Jeong, Seunghyeon Kim, Seokeon Choi, and Changick Kim.
\newblock Diversify and match: A domain adaptive representation learning
  paradigm for object detection.
\newblock In {\em Proceedings of the IEEE Conference on Computer Vision and
  Pattern Recognition}, pages 12456--12465, 2019.

\bibitem{lee2019sliced}
Chen-Yu Lee, Tanmay Batra, Mohammad~Haris Baig, and Daniel Ulbricht.
\newblock Sliced wasserstein discrepancy for unsupervised domain adaptation.
\newblock In {\em Proceedings of the IEEE Conference on Computer Vision and
  Pattern Recognition}, pages 10285--10295, 2019.

\bibitem{lee2018diverse}
Hsin-Ying Lee, Hung-Yu Tseng, Jia-Bin Huang, Maneesh Singh, and Ming-Hsuan
  Yang.
\newblock Diverse image-to-image translation via disentangled representations.
\newblock In {\em Proceedings of the European Conference on Computer Vision
  (ECCV)}, pages 35--51, 2018.

\bibitem{lin2017focal}
Tsung-Yi Lin, Priya Goyal, Ross Girshick, Kaiming He, and Piotr Doll{\'a}r.
\newblock Focal loss for dense object detection.
\newblock In {\em Proceedings of the IEEE international conference on computer
  vision}, pages 2980--2988, 2017.

\bibitem{lin2014microsoft}
Tsung-Yi Lin, Michael Maire, Serge Belongie, James Hays, Pietro Perona, Deva
  Ramanan, Piotr Doll{\'a}r, and C~Lawrence Zitnick.
\newblock Microsoft coco: Common objects in context.
\newblock In {\em European conference on computer vision}, pages 740--755.
  Springer, 2014.

\bibitem{liu2016ssd}
Wei Liu, Dragomir Anguelov, Dumitru Erhan, Christian Szegedy, Scott Reed,
  Cheng-Yang Fu, and Alexander~C Berg.
\newblock Ssd: Single shot multibox detector.
\newblock In {\em European conference on computer vision}, pages 21--37.
  Springer, 2016.

\bibitem{liu2018detach}
Yen-Cheng Liu, Yu-Ying Yeh, Tzu-Chien Fu, Sheng-De Wang, Wei-Chen Chiu, and
  Yu-Chiang Frank~Wang.
\newblock Detach and adapt: Learning cross-domain disentangled deep
  representation.
\newblock In {\em Proceedings of the IEEE Conference on Computer Vision and
  Pattern Recognition}, pages 8867--8876, 2018.

\bibitem{locatello2018challenging}
Francesco Locatello, Stefan Bauer, Mario Lucic, Sylvain Gelly, Bernhard
  Sch{\"o}lkopf, and Olivier Bachem.
\newblock Challenging common assumptions in the unsupervised learning of
  disentangled representations.
\newblock 2019.

\bibitem{pan2019transferrable}
Yingwei Pan, Ting Yao, Yehao Li, Yu Wang, Chong-Wah Ngo, and Tao Mei.
\newblock Transferrable prototypical networks for unsupervised domain
  adaptation.
\newblock In {\em Proceedings of the IEEE Conference on Computer Vision and
  Pattern Recognition}, pages 2239--2247, 2019.

\bibitem{peng2019domain}
Xingchao Peng, Zijun Huang, Ximeng Sun, and Kate Saenko.
\newblock Domain agnostic learning with disentangled representations.
\newblock {\em ICML}, 2019.

\bibitem{redmon2016you}
Joseph Redmon, Santosh Divvala, Ross Girshick, and Ali Farhadi.
\newblock You only look once: Unified, real-time object detection.
\newblock In {\em Proceedings of the IEEE conference on computer vision and
  pattern recognition}, pages 779--788, 2016.

\bibitem{ren2015faster}
Shaoqing Ren, Kaiming He, Ross Girshick, and Jian Sun.
\newblock Faster r-cnn: Towards real-time object detection with region proposal
  networks.
\newblock In {\em Advances in neural information processing systems}, pages
  91--99, 2015.

\bibitem{ridgeway2018learning}
Karl Ridgeway and Michael~C Mozer.
\newblock Learning deep disentangled embeddings with the f-statistic loss.
\newblock In {\em Advances in Neural Information Processing Systems}, pages
  185--194, 2018.

\bibitem{roy2019unsupervised}
Subhankar Roy, Aliaksandr Siarohin, Enver Sangineto, Samuel~Rota Bulo, Nicu
  Sebe, and Elisa Ricci.
\newblock Unsupervised domain adaptation using feature-whitening and consensus
  loss.
\newblock In {\em Proceedings of the IEEE Conference on Computer Vision and
  Pattern Recognition}, pages 9471--9480, 2019.

\bibitem{russo2018source}
Paolo Russo, Fabio~M Carlucci, Tatiana Tommasi, and Barbara Caputo.
\newblock From source to target and back: symmetric bi-directional adaptive
  gan.
\newblock In {\em Proceedings of the IEEE Conference on Computer Vision and
  Pattern Recognition}, pages 8099--8108, 2018.

\bibitem{saito2019strong}
Kuniaki Saito, Yoshitaka Ushiku, Tatsuya Harada, and Kate Saenko.
\newblock Strong-weak distribution alignment for adaptive object detection.
\newblock In {\em Proceedings of the IEEE Conference on Computer Vision and
  Pattern Recognition}, pages 6956--6965, 2019.

\bibitem{sakaridis2018semantic}
Christos Sakaridis, Dengxin Dai, and Luc Van~Gool.
\newblock Semantic foggy scene understanding with synthetic data.
\newblock {\em International Journal of Computer Vision}, 126(9):973--992,
  2018.

\bibitem{scott2018adapted}
Tyler Scott, Karl Ridgeway, and Michael~C Mozer.
\newblock Adapted deep embeddings: A synthesis of methods for k-shot inductive
  transfer learning.
\newblock In {\em Advances in Neural Information Processing Systems}, pages
  76--85, 2018.

\bibitem{Sutskever2013On}
I. Sutskever, J. Martens, G. Dahl, and G. Hinton.
\newblock On the importance of initialization and momentum in deep learning.
\newblock In {\em International Conference on International Conference on
  Machine Learning}, 2013.

\bibitem{szegedy2016rethinking}
Christian Szegedy, Vincent Vanhoucke, Sergey Ioffe, Jon Shlens, and Zbigniew
  Wojna.
\newblock Rethinking the inception architecture for computer vision.
\newblock In {\em Proceedings of the IEEE conference on computer vision and
  pattern recognition}, pages 2818--2826, 2016.

\bibitem{vu2019cascade}
Thang Vu, Hyunjun Jang, Trung~X Pham, and Chang~D Yoo.
\newblock Cascade rpn: Delving into high-quality region proposal network with
  adaptive convolution.
\newblock {\em arXiv preprint arXiv:1909.06720}, 2019.

\bibitem{wang2019few}
Tao Wang, Xiaopeng Zhang, Li Yuan, and Jiashi Feng.
\newblock Few-shot adaptive faster r-cnn.
\newblock In {\em Proceedings of the IEEE Conference on Computer Vision and
  Pattern Recognition}, pages 7173--7182, 2019.

\bibitem{xie2019multi}
Rongchang Xie, Fei Yu, Jiachao Wang, Yizhou Wang, and Li Zhang.
\newblock Multi-level domain adaptive learning for cross-domain detection.
\newblock {\em arXiv preprint arXiv:1907.11484}, 2019.

\bibitem{zhang2019domain}
Yabin Zhang, Hui Tang, Kui Jia, and Mingkui Tan.
\newblock Domain-symmetric networks for adversarial domain adaptation.
\newblock In {\em Proceedings of the IEEE Conference on Computer Vision and
  Pattern Recognition}, pages 5031--5040, 2019.

\bibitem{zhu2019adapting}
Xinge Zhu, Jiangmiao Pang, Ceyuan Yang, Jianping Shi, and Dahua Lin.
\newblock Adapting object detectors via selective cross-domain alignment.
\newblock In {\em Proceedings of the IEEE Conference on Computer Vision and
  Pattern Recognition}, pages 687--696, 2019.

\end{thebibliography}
}


\end{document}